\documentclass[a4paper, 10pt, conference]{ieeeconf}      

\usepackage{MyStyle}

\FGfinalcopy

\IEEEoverridecommandlockouts                              

\usepackage{graphics}
\usepackage{epsfig} 
\usepackage{mathptmx} 
\usepackage{times}
\usepackage{amsmath} 
\usepackage{amssymb}  
\usepackage{caption}
\usepackage{tabularx}
\usepackage{array}

\usepackage{booktabs}

\title{\LARGE \bf
3DRealHead: Few-Shot Detailed Head Avatar
}

\author{\parbox{16cm}{\centering
    {\large Jalees Nehvi$^1$, Timo Bolkart$^2$, Thabo Beeler$^2$, Justus Thies$^1$}\\
    {\normalsize
    $^1$ Technical University of Darmstadt\\
    $^2$ Google}}
    \thanks{Project Page:}
}

\newcolumntype{Y}{>{\centering\arraybackslash}X}

\usepackage{hyperref}
\makeatletter
\let\NAT@parse\undefined
\usepackage[numbers,sort&compress]{natbib}
\makeatother
\begin{document}
\thispagestyle{plain}
\pagestyle{plain}

\pagenumbering{arabic}
\makeatletter
\def\@oddfoot{\hfil\thepage\hfil}
\def\@evenfoot{\hfil\thepage\hfil}
\makeatother

\ifFGfinal
\thispagestyle{empty}
\pagestyle{empty}
\else
\pagestyle{plain}
\fi

\twocolumn[{%
\renewcommand\twocolumn[1][]{#1}%
\maketitle
\begin{center}
    \centering
    \vspace{-0.7cm}
    \includegraphics[clip, trim=0cm 5.1cm 0cm 3cm, width=0.85\linewidth]{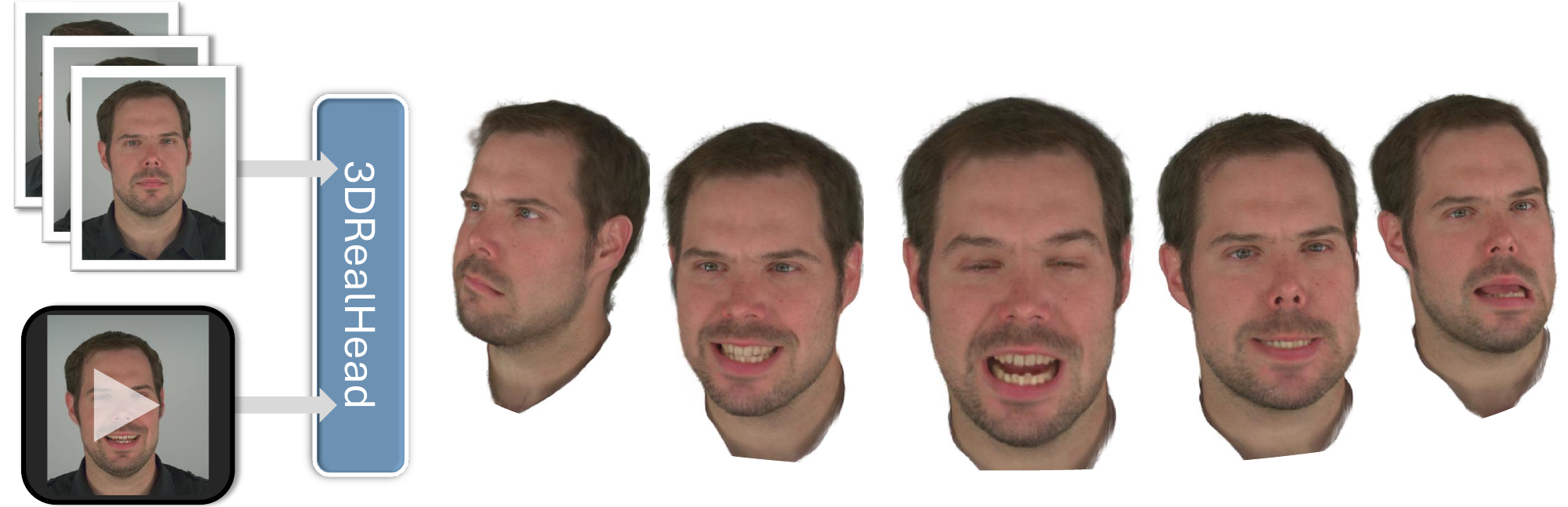}
    \captionof{figure}{
    Given a few-shot input (1-3 images), we reconstruct a 3D head avatar that can be driven by a monocular input video.
    We leverage both compressed 3DMM facial expression conditioning as well as detailed features of the mouth region in the input video to recover a detailed 3D head avatar based on 3D Gaussian primitives.
    }
    \vspace{0.2cm}
    \label{fig:teaser}
\end{center}%
}]
\thispagestyle{plain}
\pagestyle{plain}

\begin{abstract}
The human face is central to communication.
For immersive applications, the digital presence of a person should mirror the physical reality, capturing the users idiosyncrasies and detailed facial expressions.
However, current 3D head avatar methods often struggle to faithfully reproduce the identity and facial expressions, despite having multi-view data or learned priors.
Learning priors that capture the diversity of human appearances, especially, for regions with highly person-specific features, like the mouth and teeth region is challenging as the underlying training data is limited.
In addition, many of the avatar methods are purely relying on 3D morphable model-based expression control which strongly limits expressivity.
To address these challenges, we are introducing 3DRealHead, a few-shot head avatar reconstruction method with a novel expression control signal that is extracted from a monocular video stream of the subject.
Specifically, the subject can take a few pictures of themselves, recover a 3D head avatar and drive it with a consumer-level webcam.
The avatar reconstruction is enabled via a novel few-shot inversion process of a 3D human head prior which is represented as a Style U-Net that emits 3D Gaussian primitives which can be rendered under novel views.
The prior is learned on the NeRSemble dataset.
For animating the avatar, the U-Net is conditioned on 3DMM-based facial expression signals, as well as features of the mouth region extracted from the driving video.
These additional mouth features allow us to recover facial expressions that cannot be represented by the 3DMM leading to a higher expressivity and closer resemblance to the physical reality. 
We compare our proposed method against one-shot prior-based and state-of-the-art monocular avatar approaches showing that our approach achieves competitive visual quality and expressive mouth articulation while requiring only a few input frames, in contrast to video-based methods that rely on dense per-identity training data.
Our project page: \href{https://jalees018.github.io/3DRealHead/}{\textit{https://jalees018.github.io/3DRealHead}}.

\end{abstract}

\section{Introduction}

The human face plays a pivotal role in communication.
In immersive applications, a person’s digital presence should closely reflect their physical reality~—~capturing individual nuances and detailed facial expressions.
Yet, existing 3D head avatar methods often fall short in faithfully reproducing identity and expression, even when equipped with multi-view data or learned priors.
One of the key challenges lies in modeling regions with highly person-specific features, such as the mouth and teeth.
In observations, these regions are often not fully visible, and thus, hard to reconstruct.
Priors could be employed to mitigate the missing data, however, since the region is very person specific and unique, there will be a loss of identity information.
Moreover, many current approaches rely solely on 3D morphable model (3DMM)-based expression control~\cite{qian2024gaussianavatars,Zielonka2022InstantVH,zielonka2025synshot}, which inherently restricts expressivity to the 3DMM expression space.
A few other approaches~\cite{latentAvatar,deep_appearance_models,saito2024rgca} extract `codecs' from an input signal to drive an avatar using a conditional variational auto-encoder (C-VAE).
These `codecs' are rich in representing the facial expression and its appearance, however, require a large dataset corpus to train the C-VAE.

We are following a similar idea and introduce 3DRealHead, a few-shot head avatar reconstruction method that leverages a hybrid expression control signal extracted from a monocular video stream, consisting of 3DMM expressions and appearance features.
With just a few images, users can generate a personalized 3D head avatar and animate it using a standard webcam.
Our method employs a unique few-shot inversion process of a 3D human head prior, represented as a Style U-Net~\cite{wang2023styleavatar} that outputs 3D Gaussian primitives renderable from novel viewpoints.
This prior is trained on the NeRSemble dataset~\cite{kirschstein2023nersemble} with additional adversarial losses.
For animation, the U-Net is conditioned not only on 3DMM-based expression signals but also on mouth-region features extracted from the driving video.
These additional features enable the avatar to express facial dynamics beyond the capabilities of traditional 3DMMs, resulting in more lifelike and expressive outputs.
We benchmark 3DRealHead against both few-shot prior-based methods and state-of-the-art monocular avatar techniques.
Our approach consistently delivers sharper visual fidelity and significantly improved expressiveness in the mouth region, bringing digital avatars closer to the realism of their physical counterparts.

\medskip
\noindent
In summary, we list the following key contributions:
\begin{itemize}
    \item a novel human head prior based on 3D Gaussian primitives that can be used to recover a 3D head from only 1-3 input images.
    \item a novel expression conditioning scheme, which allows us to extract coarse facial expressions (3DMM-based) and fine-scale mouth appearance of the driving video, and apply it to our reconstructed 3D head avatar.
\end{itemize}

\begin{figure*}
    \hspace{3mm}
    \includegraphics[clip, trim=0cm 6.6cm 0cm 2.75cm, width=\linewidth]{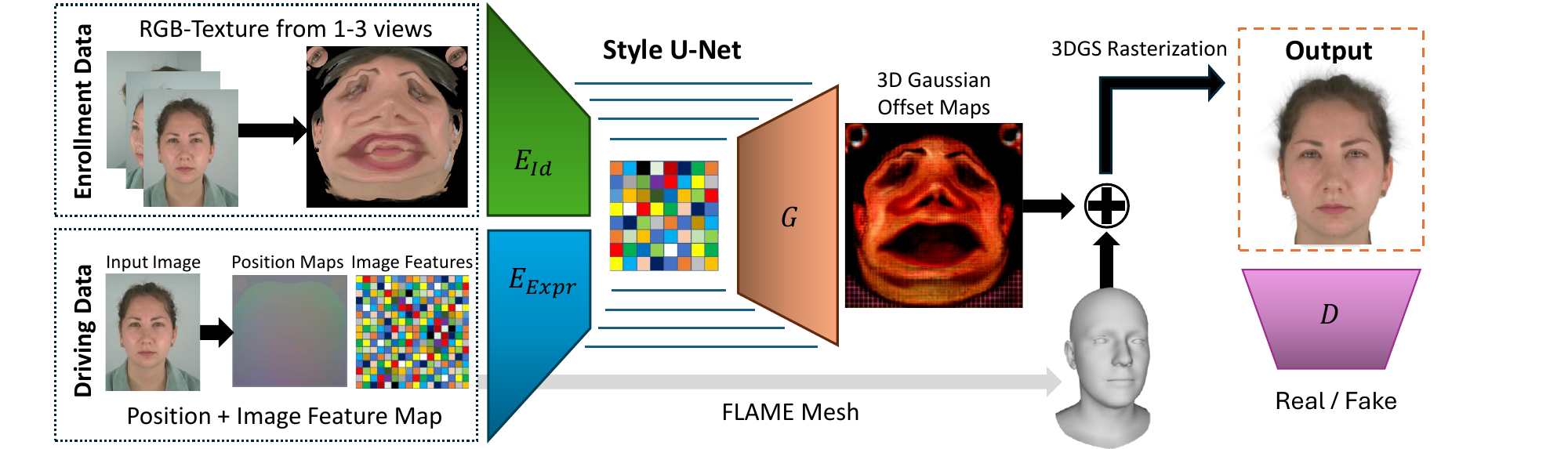}
    \vspace{2mm}
    \caption{\textbf{Overview of 3DRealHead.} Based on few-shot input data, an RGB texture is generated which serves as identity conditioning to the Style U-Net which predicts 3D Gaussian primitives in UV-space relative to a FLAME head mesh. To control detailed facial expressions from a driving input video, we extract 3DMM-based position maps~\cite{SMIRK:CVPR:2024} and image gradient features which are projected to the UV texture space. From this control signal, the model learns how to produce dynamically changing 3D Gaussian primitives which can be rendered using 3D Gaussian splatting. The Style U-Net is trained on the NeRSemble~\cite{kirschstein2023nersemble} dataset, such that it can act as a prior for few-shot inversion, where only the identity encoder $E_{Id}$ is finetuned. During the training of the prior, we employ several reconstruction losses, as well as a multi-head discriminator $D$ for adversarial training.
    }
    \label{fig:pipeline}
\end{figure*}

\section{Related Work}

The related work can be reviewed across the axis of required data to recover a 3D avatar.
Avatars with a very high appearance quality are typically reconstructed with data captured in a multi-view studio setup.
As this type of data is not widely available, especially not in the context of a general user at home, a diverse set of methods has been developed that either take monocular video or even just a few photographs of a subject as input.
Our method also aims at recovering a 3D avatar from a few images, followed by driving it with a monocular input stream where the motion can be extracted from.
%

\medskip\noindent
\textbf{3D head avatars from multi-view data.}
With the release of a series of mutli-view datasets, like
NeRSemble~\cite{kirschstein2023nersemble}, Multiface~\cite{wuu2022multiface}, and Ava-256~\cite{NEURIPS2024_9712b783}, a number of head avatar reconstruction methods that are specialized to use such high quality data were introduced.
The initial set of works concentrated on building personalized avatars from rich data recordings of a subject.
Gaussian Avatars~\cite{qian2024gaussianavatars} is one of the first methods leveraging the NeRSemble dataset to demonstrate how 3D Gaussian primitives attached to a 3DMM (FLAME~\cite{Li2017flame}) can be reconstructed from calibrated multi-view data.
Other works like Gaussian Head Avatar~\cite{xu2023gaussianheadavatar}, GaussianHeads~\cite{teotia24gaussianheads} or D3GA~\cite{zielonka25dega} followed.
In GEM~\cite{zielonka2024gem}, a method to compress such a representation has been demonstrated that enables the avatar to be displayed and animated in real-time.
While these methods based on 3D Gaussian primitives are very closely related to our method which uses the same representation, there are also NeRF-based works~\cite{kabadayi24ganavatar} or mixture of volumetric primitives~\cite{Lombardi21} that use multi-view data.
Note that our approach relies on a prior that is trained on multi-view data, however, during test time it
does not rely on multi-view data.
A few images of a subject with neutral expression as input are sufficient to recover an animatable 3D avatar.
A monocular video is then used to drive the avatar.
%

\medskip\noindent
\textbf{3D head avatars from monocular video data.}
In contrast to the multi-view data, methods for monocular video data are heavily based on priors.
The most common prior is a 3D morphable model (3DMM)~\cite{10.1145/311535.311556,Li2017flame,egger20203dmorphablefacemodels} which reduces the dimensionality of the `face space' (geometry and appearance) to a low dimensional space.
Works like Face2Face~\cite{thies2016face} showed impressive results with such mesh-based representations.
With the advances in neural rendering, methods  like \cite{thies2019deferred} were proposed that employed a CNN-based refinement of the results in 2D.
%
3D neural rendering methods like NeRF~\cite{mildenhall2020nerf} were added as an additional layer on top of the 3DMM-based meshes.
Early implementations like NeRFace~\cite{gafni2020dynamicneuralradiancefields}, have been improved by integrating neural graphic primitives~\cite{mueller2022instant} which resulted in methods like INSTA~\cite{Zielonka2022InstantVH} that are able to reconstruct an avatar within a few minutes instead of hours.
With the introduction of 3D Gaussian Splatting~\cite{kerbl3Dgaussians}, 3D Gaussian primitives have been added on top of 3DMM-based meshes.
While this has been done first on multi-view data, recently, methods have been shown that work on monocular data like MonoGaussianAvatar~\cite{Chen2023MonoGaussianAvatarMG} or FlashAvatar~\cite{xiang2024flashavatar}.
A key evolution step is the usage of learned priors that support the reconstruction of such Gaussian-based avatars from monocular data.
Many of the methods are trained such that they can even work with only a few images which we will describe in the following.

\medskip\noindent
\textbf{3D head avatars from few-shot images.}
As mentioned above, methods that rely only on monocular data and especially few-shot images are heavily relying on learned priors.
These priors can be learned from (multi-view) data of real humans, or synthetic data generated in modern rendering frameworks.
A major limitation is the restricted variety of subjects that one has in studio-captured datasets.
As a result, there are methods~\cite{Chan2021,an2023panohead,kirschstein2024gghead} that learn based on `in-the-wild' data (photo-collections from the internet) to recover a 3D prior.
However, those priors are often limited in their facial expressions and their view-stability and quality.
A different line of work focuses on using synthetic datasets to learn a prior like Cafca~\cite{buehler2024cafca}, Rodin~\cite{wang2023rodin}, and RodinHD~\cite{zhang2025rodinhd}.
Similarly, in SynShot~\cite{zielonka2025synshot}, they build a synthetic prior, but then show how it can be finetuned on few-shot input data to adapt to a real subject.
In HeadGAP~\cite{zheng2025headgapfewshot3dhead}, PreFace~\cite{buhler2023preface}, One2Avatar~\cite{yu2024one2avatargenerativeimplicithead}, similar schemes are presented that operate on datasets of real subjects.
Different to our method, there are also methods that use a 2D diffusion prior~\cite{kirschstein2024diffusionavatars,chen2024morphable} or 3D diffusion~\cite{lan2023gaussian3diff} to recover an avatar.

In our method, we are following a similar approach as SynShot and train on NeRSemble~\cite{kirschstein2023nersemble}.
We show that despite the limited diversity of the real data, it is possible to adapt and recover the appearance of a subject from a few input views.
In contrast to SynShot, which is restricted by the underlying 3DMM, we propose a novel architecture that allows for additional appearance features as input that widens the space of facial expressions, especially, the mouth interior.

\section{Prerequisites}

\noindent
\textit{3D Gaussian Representation.}
3D Gaussian Splatting (3DGS)~\cite{kerbl3Dgaussians} is a recent splatting-based fast rasterization technique that treats 3D objects or scenes as a set of 3D Gaussian ellipsoids, each with its mean $\mu$ representing the ellipsoid position and its covariance matrix $\Sigma$ controlling the 3D scale and orientation.
The opacity $\alpha$ controls the visibility of the ellipsoids and the Spherical Harmonic coefficients $c$ model the view-dependent color.
Each 3D Gaussian primitive has its probability density function defined as:
\begin{equation}
p(\mathbf{x}) = \frac{1}{ (2 \pi)^{3/2} \, |\Sigma|^{1/2} } 
\exp\!\left( -\tfrac{1}{2} (\mathbf{x} - \mu)^\top \Sigma^{-1} (\mathbf{x} - \mu) \right) .
\end{equation}
Each 3D Gaussian is projected onto the 2D image plane as an ellipse, followed by blending the opacity $\alpha$ of overlapping Gaussians resulting in the final color $c$. Given a set of images and corresponding camera pose and intrinsics, differentiable 3D Gaussian Splatting~\cite{kerbl3Dgaussians} can be used to reconstruct 3D Gaussian attributes, and it allows for real-time rendering under novel views.

\medskip
\noindent
\textit{FLAME Model.} We use FLAME~\cite{Li2017flame} as our underlying 3D head morphable model.
FLAME is a statistical model parametrized by the facial identity/shape $\beta$, expressions $\psi$, and pose $\theta$ representing rotations of the jaw, neck and head regions.
Given the template $\bar{T}$, the resulting vertices $\mathbb{M}_C$ in canonical space is a function of $\beta$,$\psi$ and $\theta$ defined as:
\begin{equation}
\mathbb{M}_C(\boldsymbol{\beta}, \boldsymbol{\psi}, \boldsymbol{\theta} ) 
= \bar{T} \;+\; B_S(\boldsymbol{\beta}) \;+\; B_E(\boldsymbol{\psi}) \;+\; B_P(\boldsymbol{\theta}),
\end{equation}
where $B_S$, $B_E$ and $B_P$ are the shape, expression and pose blendshapes respectively.
Using linear blend skinning (LBS)~\cite{10.1145/2659467.2675048}, the mesh $\mathbb{M}_C$ is then transformed into the final mesh $\mathbb{M}$:
\begin{equation}
\mathbb{M} \;=\;
LBS\!\left( \mathbb{M}_C, \; J(\boldsymbol{\beta}), \; \boldsymbol{\theta}, \; \mathcal{W} \right),
\end{equation}
where $\mathcal{W}$ are the FLAME model skinning weights, and $J$ is the joint regressor of FLAME.
We use VHAP~\cite{qian2024vhap} and SMIRK~\cite{SMIRK:CVPR:2024} to extract the FLAME parameters from monocular video data to drive the reconstructed avatar.

\section{Method}

Given a few images (1-3) of the subject in neutral facial expression, our method \textit{3DRealHead} reconstructs a personalized 3D head avatar that can be rendered under novel views and expressions, driven by a monocular video.
Our method employs a Style U-Net \cite{wang2023styleavatar} architecture conditioned on identity and expression maps, that generates 3D Gaussian primitives as the primary 3D representation, which are then offset on top of a canonical 3D morphable model (3DMM) \cite{Li2017flame} and subsequently deformed using the 3DMM expressions and pose via linear blend skinning. The deformed 3D Gaussians are finally rasterized \cite{kerbl3Dgaussians} along novel views.

\subsection{Model Backbone}

\medskip
We choose the Style U-Net \cite{wang2023styleavatar} architecture with skip connections as our model backbone.
Our model consists of two CNN-based encoders encoding the identity and expression information separately.
Both encoders take as input identity and expression information in the UV space.
Specifically, the identity encoder $E_{id} (M_{id})$ takes as input neutral identity texture maps $ M_{id} \in \mathbb{R}^{H \times W \times 3}$ with height $H$ and width $W$.
The RGB texture maps are defined in a fixed UV space and can be partial or complete corresponding to a single view or three views respectively.
These maps represent the generic identity (shape and appearance) of the subject and contain no information about the mouth region since the said region is masked out.
The expression encoder $E_{exp} (\Delta p, M_{mouth})$ operates on two different types of conditioning signals.
Firstly, the position offsets $\Delta p \in \mathbb{R}^{H \times W \times 3}$ obtained from SMIRK \cite{SMIRK:CVPR:2024}. Here, we use only the FLAME expressions and jaw pose parameters.
The shape and pose parameters are set to their default canonical values.
The FLAME positions corresponding to these regressed parameters are finally subtracted from the neutral frame positions belonging to the same identity to obtain expression position offsets in canonical space resulting in $\Delta p$, thereby disentangling shape information from the expressions.
Secondly, for the mouth region, we introduce additional conditionings in the form of image gradient maps $ M_{mouth} \in \mathbb{R}^{H \times W \times 3}$ in UV space that contain information about the structure of the mouth interior (other regions are set to zero).
Both the expression as well as the mouth region conditionings are concatenated and fed to the expression encoder.

The resulting identity and expression features are combined and passed through a common decoder $G(E_{id} (M_{id}),E_{exp} (\Delta p, M_{mouth}))$ via multiple skip connections at different levels.
This ensures that the input features at multiple resolutions representing varying scales of information, both coarse and fine, are fed into the decoder.

The skip connections as well as the output from previous layers are decoded by the up-convolution blocks.
The decoder finally predicts $n$ 3D Gaussian primitives $\Delta G = \{\Delta x, \alpha, \omega, s, c\}$ in the canonical UV texture space, where $\Delta x \in \mathbb{R}^{n \times 3}$ is the position offset, $\alpha \in \mathbb{R}^{n}$ the opacity, $\omega \in \mathbb{R}^{n \times 4}$ the quaternion orientation, $s \in \mathbb{R}^{n \times 3}$ the scale and $c \in \mathbb{R}^{n \times 16 \times 3}$ the spherical harmonic coefficients of degree three modeling the color of each 3D Gaussian.

The position offsets are added to the FLAME vertices in canonical space and linear-blend-skinning (LBS) \cite{10.1145/2659467.2675048} is applied resulting in a final 3D Gaussian position $X$:
\begin{equation}
X \;=\;
LBS\!\left(
\mathbb{M}_C(\boldsymbol{\beta}, \boldsymbol{\psi}, \boldsymbol{\theta} )
 + \Delta x,\; J(\boldsymbol{\beta}), \; \boldsymbol{\theta}, \; \mathcal{W} \right). 
\end{equation}
Note that the covariance matrices of the 3D Gaussians (which are represented by the quaternion $\omega$ and a scale $s$) are transformed accordingly with the linear blendskinning transformation resulting in a final covariance $\Sigma$.
Using the 3D Gaussians defined by $X$ and $\Sigma$, we employ 3D Gaussian splatting to compute the output image $I$ in a differentiable manner.

\subsection{Training} We train our model on multi-view images in a supervised manner.
Since multi-view high quality data is limited and in order to generalize better to In-the-Wild data, we apply random augmentations to the RGB input conditionings and the corresponding ground truth images.
We introduce color jitter to the training images with $50\%$ probability by randomly varying the hue with factor 0.1 and contrast, intensity, and saturation values with factor 0.3, in addition to applying random gaussian blur with kernel size 5 and std. deviation in $[0.1, 1]$. 
We introduce adversarial training \cite{goodfellow2014generativeadversarialnetworks} in the later stages.
We use a transformer-based multi-head discriminator similar to StyleGAN-T \cite{Sauer2023ICML} that operates on DINO \cite{caron2021emerging} features.
The classification is performed using 5 similar discriminator heads, each operating on different levels of the DINO token features.
We minimize the following objective:
\begin{equation}
\mathcal{L}(E_{id},E_{exp},G) = \, \mathcal{L}_{\text{photo}} 
                             + \lambda_{pos} \, \mathcal{L}_{\text{pos}} 
                             + \lambda_{scal} \, \mathcal{L}_{\text{scal}},
                             + \lambda_{gan} \, \mathcal{L}_{\text{gan}},
                             \label{eq:Train1}
\end{equation}
\begin{equation}
\mathcal{L}_{\text{photo}} = \lambda_{L1} \, \mathcal{L}_{\text{L1}} 
                             + \lambda_{VGG} \, \mathcal{L}_{\text{VGG}} 
                             + \lambda_{SSIM} \, \mathcal{L}_{\text{SSIM}},
                             \label{eq:Train2}
\end{equation}
where $\mathcal{L}_{\text{L1}}$ is the $L_{1}$ photometric loss, $\mathcal{L}_{\text{VGG}}$ is the perceptual loss \cite{johnson2016perceptuallossesrealtimestyle} and $\mathcal{L}_{\text{SSIM}}$ is the structural similarity (SSIM) \cite{wang2004image} loss between the rendered and ground truth images.
$\mathcal{L}_{\text{pos}}$ and $\mathcal{L}_{\text{scal}}$ are the $L_{1}$  position and scale regularization on the predicted position and scale offset primitives, respectively, and $\mathcal{L}_{\text{gan}}$ is the adversarial loss \cite{goodfellow2014generativeadversarialnetworks}.
\subsection{Few-shot Inference}
The inference stage takes as input 1-3 images of the test subject along with the driving signals and generates a 3D head avatar that can be rasterized using novel pose or camera view to obtain the final image.
We primarily divide the few-shot inference process into two stages:
\subsubsection{Enrollment}
During this stage, we capture our test person under neutral expression along the front or sides additionally to obtain 1-3 images of that subject.
We then perform inversion where we reconstruct the 3D head avatar by optimizing for the identity encoder $E_{id}$ output namely the latent feature skip maps $F=\{F_{i}\}_{i=1}^{K}$ where $F_{i}=E_{id}(M_{id})$ along with the final layer of the decoder $G$. Optimizing for multiple multi-resolution identity feature maps instead of a single latent code vector results in sharper 3D avatar quality.
We freeze the rest of the model and do not fine-tune the remaining part, this ensures that the prior information in the decoder is preserved.  We use the captured images as targets for supervision by minimizing the following objective:
\begin{equation}
\mathcal{L}(F) = \, \mathcal{L}_{\text{photo}} 
                             + \lambda_{pos} \, \mathcal{L}_{\text{pos}} 
                             + \lambda_{scal} \, \mathcal{L}_{\text{scal}}
                             + \lambda_{GReg} \, \mathcal{L}_{\text{GReg}},
                             \label{eq:Inv1}
\end{equation}
\begin{equation}
\mathcal{L}_{\text{GReg}}=\|G(E_{id},E_{exp}) - G(F^{*},E_{exp}) \|_{1}.
\end{equation}
where $\mathcal{L}_{\text{photo}}$ is the same as \eqref{eq:Train2}. We introduce an additional $L_{1}$ regularization loss $\mathcal{L}_{\text{GReg}}$ where we force our optimized primitives to be as close as possible to the initial 3D primitives. This prevents overfitting to the target views. The entire inversion process takes around 2-3 minutes for 3 target frames. For further details, please refer to Appendix \ref{appendix:Inference}.
\subsubsection{Driving}
Once we obtain the optimized identity features $F^{*}$ for a given target identity, we drive the resulting 3D Gaussians via a monocular video.
We track the sequence using VHAP tracker \cite{qian2024vhap} and obtain the FLAME parameters for driving our avatar. We also run the SMIRK regressor to obtain expression position map and use the VHAP tracked meshes to obtain the mouth conditioning map as inputs to the expression encoder.
The decoder takes in both the expression encoder output as well as the optimized identity maps and uses the FLAME trackings to drive the resulting 3D Gaussians representing the 3D head avatar, for each input frame which are finally rendered along a novel view.
The entire driving plus rendering stage runs at 27 fps.

\subsection{Implementation Details.} We employ the original 3DGS~\cite{kerbl3Dgaussians} PyTorch version as rasterizer. We use PyTorch Adam \cite{Kingma2014AdamAM} optimizer with a learning rate of $2.5e^{-4}$ which is decayed progressively. We train our model on a single 48 GB NVIDIA L40S GPU with a batch size of 2 for 1,000,000 iterations which takes around 5 days.
We use a resolution of $512 \times 512$ for both the encoder input conditioning maps as well as the 3D Gaussian offset map obtained from the decoder output.
In \eqref{eq:Train1}, \eqref{eq:Train2} and \eqref{eq:Inv1}, we use the loss term weights: $\lambda_{pos}=1$, $\lambda_{scal}=0.1$, $\lambda_{L1}=5$, $\lambda_{VGG}=0.1$, $\lambda_{SSIM}=0.2$, $\lambda_{gan}=0.01$ and $\lambda_{GReg}=5$.

\section{Results}


In this section, we discuss the effectiveness of our method in comparison to existing works based on high quality studio data as well as In-the-Wild monocular data, and justify our design choices with ablation studies.
\begin{table}[t]
    \vspace{-5mm}
    \caption{Quantitative comparison of our method against prior-based and SOTA monocular methods on Studio and In-the-Wild datasets (8 subjects). For studio data, we perform 1-shot inversion.}
    \centering
    \small
    \resizebox{\linewidth}{!}{
    \begin{tabular}{|c|c|c|c|c|c|}
        \hline
        \multicolumn{6}{|c|}{\textbf{Studio Data}} \\
        \hline
        \hline
        \textbf{Method} & \textbf{LPIPS} $\downarrow$ & \textbf{SSIM} $\uparrow$ & \textbf{PSNR} $\uparrow$ & \textbf{ID} $\uparrow$ & \textbf{t-LPIPS} $\downarrow$ \\
        \hline
        Ours & \textbf{0.036} & \textbf{0.944} & \textbf{33.152} & \textbf{0.992} & \textbf{0.073} \\
        GAGAvatar \cite{chu2024gagavatar} & 0.037 & 0.935 & 31.595 & \textbf{0.992} &  0.077 \\
        GPAvatar \cite{chu2024gpavatargeneralizableprecisehead} & 0.063 & 0.927 & 32.379 & 0.982 & 0.101 \\
        LAM \cite{he2025lamlargeavatarmodel} & 0.081 & 0.899 & 25.035 & 0.974 & 0.098 \\
        \hline
        \hline
        \multicolumn{6}{|c|}{\textbf{In-the-Wild Data -- Few-Shot}} \\
        \hline
        \hline
        \textbf{Method} & \textbf{LPIPS} $\downarrow$ & \textbf{SSIM} $\uparrow$ & \textbf{PSNR} $\uparrow$ & \textbf{ID} $\uparrow$ & \textbf{t-LPIPS} $\downarrow$ \\
        \hline
        Ours & 0.052$\pm$0.002 & 0.936$\pm$0.002 & 29.803$\pm$0.195 & 0.989$\pm$0.001 & 0.077$\pm$0.003\\
        SynShot \cite{zielonka2025synshot} & 0.066 & 0.930 & 27.880 & 0.988 & 0.103 \\
        \hline
        \multicolumn{6}{|c|}{\textbf{In-the-Wild Data -- Video-based}} \\
        \hline
        INSTA \cite{Zielonka2022InstantVH} & 0.048 & 0.951 & 32.081 & 0.991 & 0.095 \\
        FlashAv. \cite{xiang2024flashavatar} & 0.050 & 0.939 & 30.184 & 0.987 & 0.093\\
        SplattingAv. \cite{shao2024splattingavatar} & 0.040 & 0.952 & 30.989 & 0.994 & 0.072 \\
        \hline
    \end{tabular}}
    \label{tab:studio_itw}
    \vspace{-3mm}
\end{table}

\subsection{Datasets}
We train our proposed method on the muti-view NeRSemble dataset~\cite{kirschstein2023nersemble}. 
Our training corpus comprises 200 real subjects captured under studio-quality conditions, each with 6 expression sequences. Each frame is captured along 16 different camera views covering the frontal hemisphere.
We evaluate our method on 20 test subjects and their expression sequences. We extract the frames from the original videos at 25 fps and downsample them by a factor of 4 resulting in a training resolution of $802 \times 550$.
We use the VHAP tracker \cite{qian2024vhap} to obtain the FLAME tracking parameters without offsets namely the shape, expressions and pose, in addition to the camera extrinsic and intrinsic parameters.
We run SMIRK~\cite{SMIRK:CVPR:2024} on frontal images by extracting only the expression and jaw pose information while setting other parameters to their default canonical values. We rasterize the resulting mesh positions in our given UV space.
We use the frontal driving image along with the meshes and camera parameters from VHAP trackings to generate gradient mouth maps in the UV space by computing image gradients from the entire expression RGB frame, followed by masking them in the UV space for the mouth region.
For the RGB identity texture, we obtain two types of input maps: one using a single frontal image and another using three images (frontal plus two side views). We randomly mix both maps for identity conditioning during training.
To evaluate on non-studio like, in-the-wild data, we use the examples from SynShot~\cite{zielonka2025synshot}.
The dataset is publicly available and released as part of INSTA \cite{Zielonka2022InstantVH}. It consists of monocular frontal talking subjects with slight side movements. The subjects are recorded using a DSLM camera.
We use 8 subjects for evaluation. We choose three frames from the training set for enrollment and drive the resulting 3D avatar using the last 600 test frames.
%


\begin{figure}[t!]
    \centering
    \vspace{-1mm}
    \includegraphics[width=\linewidth]{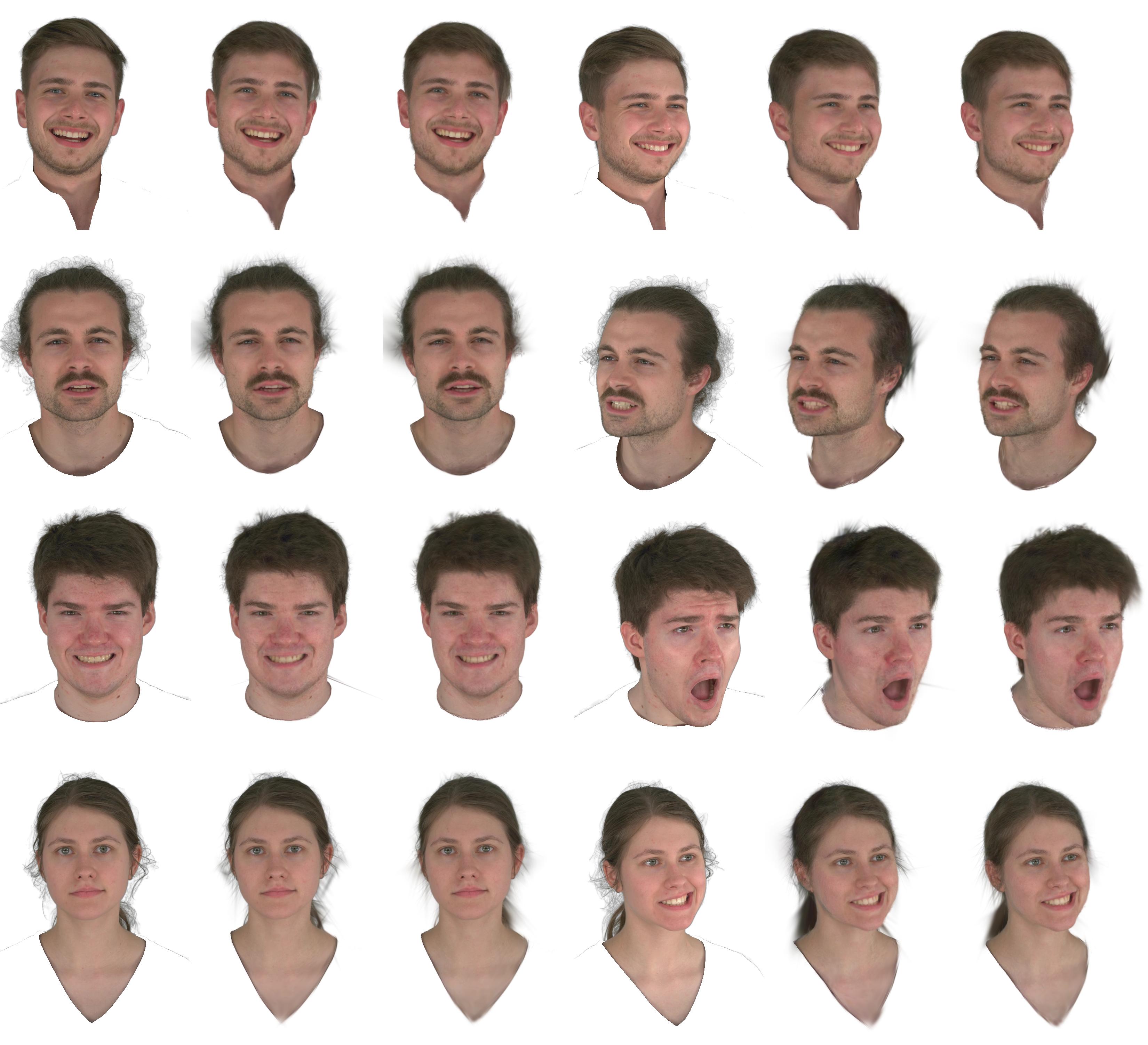}
    \vspace{2mm} 
    \begin{tabularx}{\linewidth}{YYYYYY}
        \text{GT} & \text{1-shot} & \text{3-shot} & \text{GT} & \text{1-shot} & \text{3-shot} \\
    \end{tabularx}
    \caption{We show qualitative results of our method along different views for cases with 1 and 3 target images.}
    \vspace{-4mm}
    \label{fig:Qual}
\end{figure}

\subsection{Qualitative Results}
In Fig. \ref{fig:Qual}, we show an experiment on the NeRSemble test data for 1-shot and 3-shot cases where each subject is self-driven using a random expression sequence.
We drive the subject along the frontal view and render it along a different one highlighting the mouth region quality when rendered from a different view.
We can clearly notice from the comparison that the quality of the mouth interior looks good when rendered from the sides. Moreover, for the 1-shot case, the reconstruction quality looks good along non-frontal views.
For results involving novel-view synthesis, please refer to the project page.

\subsection{Comparisons on One-Shot Studio Data}
\begin{figure*}[t!]
    \centering
    \includegraphics[width=\textwidth]{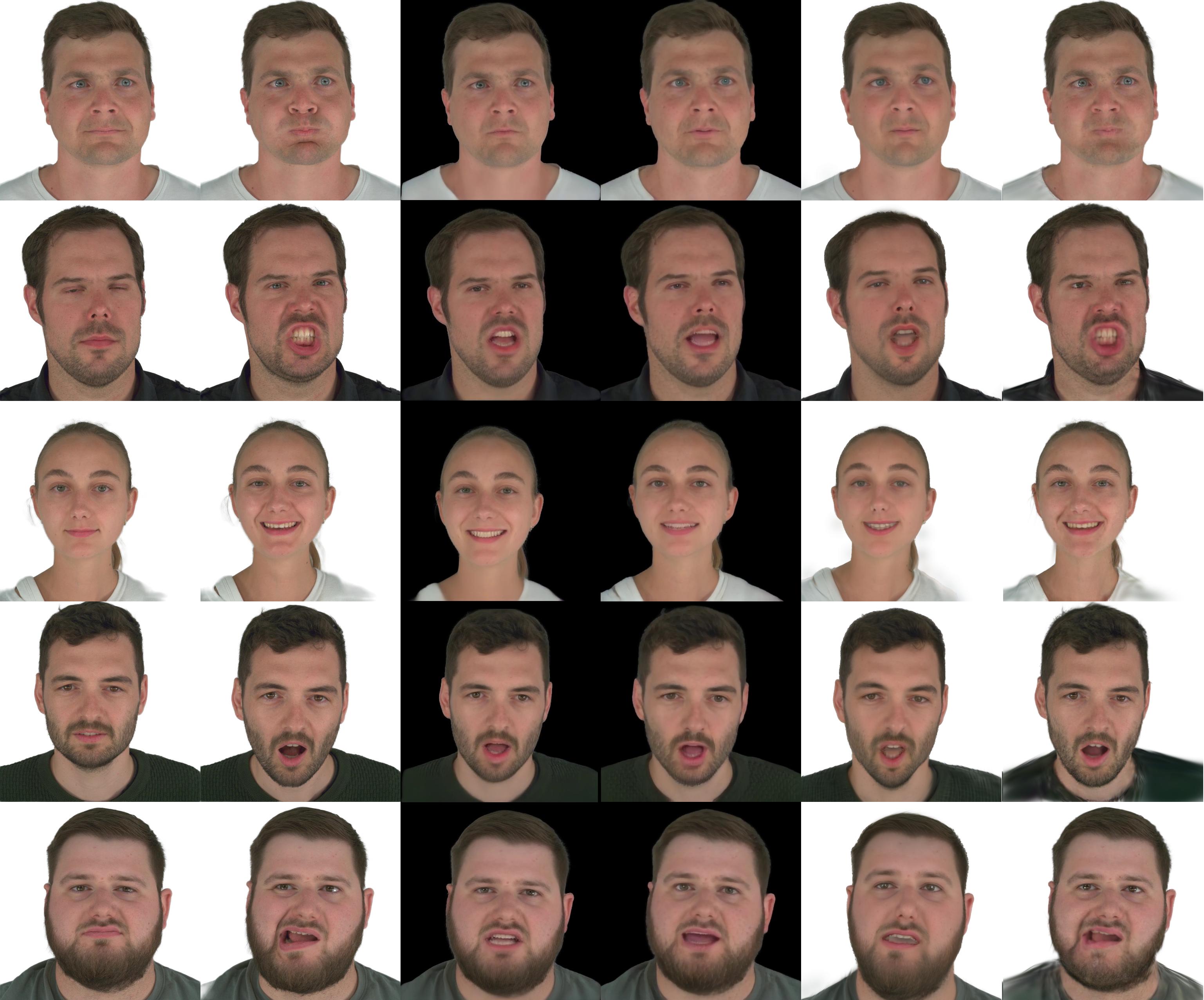}

    \vspace{2mm}
    \begin{tabularx}{\textwidth}{YYYYYY}
        Target Identity & Driving Expression & GAGAvatar \cite{chu2024gagavatar} & GPAvatar \cite{chu2024gpavatargeneralizableprecisehead} & LAM \cite{he2025lamlargeavatarmodel} & Ours
    \end{tabularx}
    
    \vspace{-1mm}
    \caption{Qualitative comparison of our approach (single target) against one-shot prior-based methods. We observe that our method produces sharper results and faithfully reproduces the driving expression.}
    \label{fig:QualComp}
    \vspace{-3mm}
\end{figure*}
We compare against one-shot prior-based methods which take monocular images as input and reconstruct an animatable avatar.
We preprocess the NeRSemble \cite{kirschstein2023nersemble} images by cropping and resizing them to 512 x 512 resolution, since the baseline methods operate directly on this resolution.
We perform a single image inversion for our method to obtain the avatar and use the same frontal frame for the competing methods.
We track the cropped dataset using the VHAP tracker and use it to drive our avatar.
We compare our method against: GAGAvatar \cite{chu2024gagavatar}, GPAvatar \cite{chu2024gpavatargeneralizableprecisehead} and LAM \cite{he2025lamlargeavatarmodel}.
GAGAvatar and GPAvatar are trained on the VFHQ \cite{xie2022vfhq} dataset which consists of frontal talking videos of subjects.
We provide the same frontal neutral image to the baselines.
We use the trackers provided with GPAvatar and GAGAvatar code-base to track the test sequence and drive the avatar.
For LAM, we use the model trained on VFHQ + NeRSemble. Since LAM also supports the VHAP tracker, we use the same tracking to drive the avatar.
We evaluate the results on 15 test subjects each associated with a random expression sequence.
In Tab. \ref{tab:studio_itw}, we report the corresponding quantitative metrics (evaluated for the head region without torso). 
From Fig. \ref{fig:QualComp} \ and Tab. \ref{tab:studio_itw}, we can observe that our method reproduces the driving expression well, while other methods struggle to give plausible expressions, especially in the mouth region.

\subsection{Comparisons on Monocular In-the-Wild Data}
\begin{figure*}[ht]
    \centering
    \vspace{-1.5mm}
    \includegraphics[width=\textwidth]{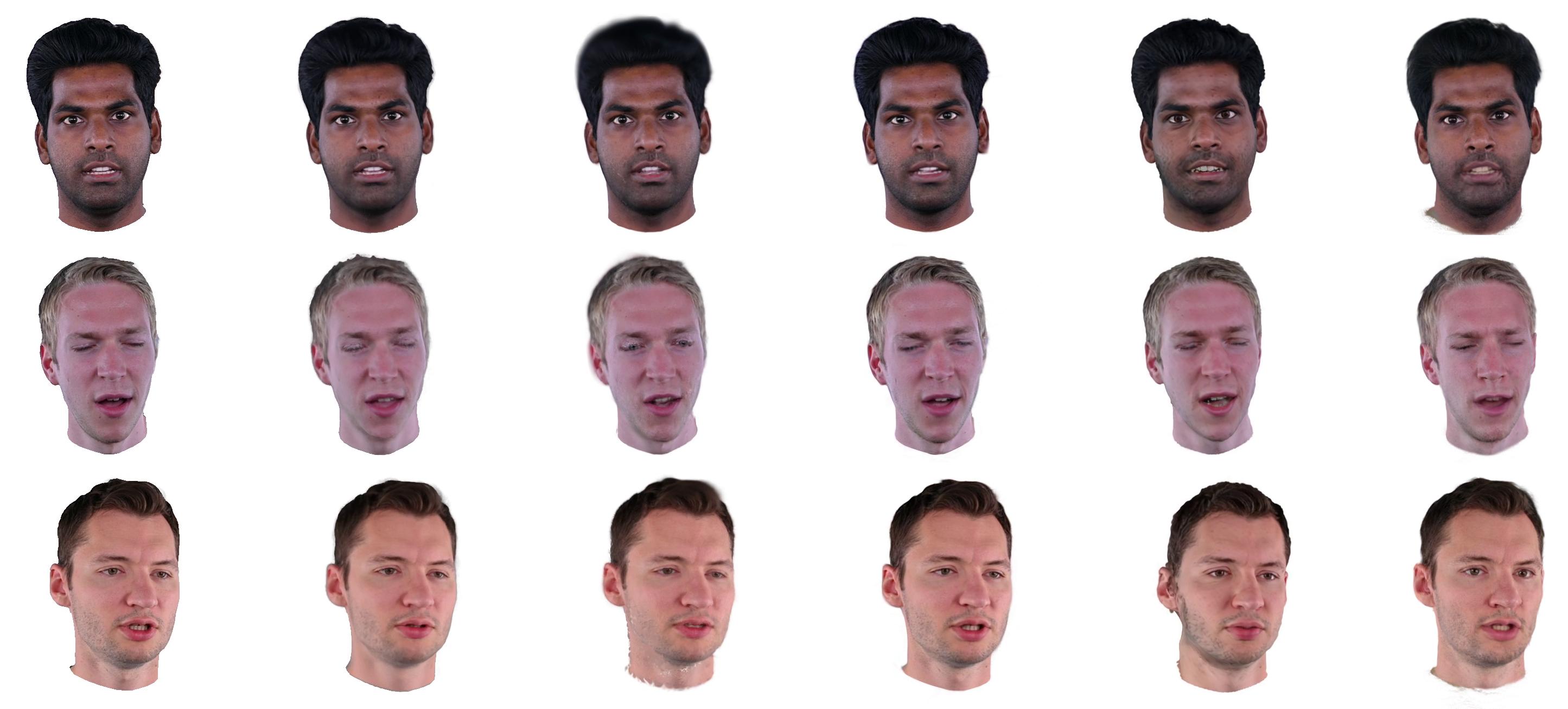}
    \vspace{1mm} 
    \begin{tabularx}{\textwidth}{YYYYYY}
        Ground Truth & INSTA \cite{Zielonka2022InstantVH} & \hspace{1mm} FlashAvatar \cite{xiang2024flashavatar} & \hspace{2mm} SplattingAvatar \hspace{4mm} \cite{shao2024splattingavatar} & \hspace{5mm} SynShot \cite{zielonka2025synshot} & \hspace{6mm} Ours
    \end{tabularx}
    \vspace{-3.5mm}
    \caption{Qualitative comparison to SynShot~\cite{zielonka2025synshot} and SOTA monocular approaches on In-the-Wild INSTA dataset~\cite{Zielonka2022InstantVH}. It is important to note that SOTA monocular methods are trained per subject on minutes of video recording while SynShot and our method are fine-tuned on only 3 frames.
    }
    \vspace{-3mm}
    \label{fig:ITW}
\end{figure*}
We compare our method against the monocular state-of-the-art methods INSTA \cite{Zielonka2022InstantVH}, FlashAvatar \cite{xiang2024flashavatar} and SplattingAvatar \cite{shao2024splattingavatar} as well as the synthetic prior-based method SynShot \cite{zielonka2025synshot}.
To create personalized avatars, the methods get videos of the INSTA dataset~\cite{Zielonka2022InstantVH} as input, comprising around 3000 mostly frontal frames for each actor.
Each avatar of the 8 subjects is then evaluated on 600 test frames.
%
Note that instead of a video, SynShot and our method perform inversion on only 3 mostly frontal frames during this evaluation.
From Fig. \ref{fig:ITW} and Tab. \ref{tab:studio_itw}, we can clearly infer that our method which is trained on limited multi-view real data and requires only 1-3 frames of the subject during inference, outperforms SynShot by a considerable margin on all metrics and is closer in most metrics to monocular video-based per-identity methods like INSTA and FlashAvatar trained on around 3000 frames per actor. However, SplattingAvatar achieves the best performance across all metrics, which can be attributed to its reliance on dense per-identity training data and strong identity-specific optimization, giving it an advantage in reconstruction fidelity. In contrast, our method is not video-based and requires only 1-3 frames during enrollment. Moreover, our approach incorporates additional mouth-region features, enabling more accurate modeling of mouth dynamics compared to SplattingAvatar.
Note that we use a single-view texture for identity conditioning in these experiments.
\subsection{Ablation Experiments}
\begin{figure}[t!]
    \centering
    \vspace{-3.5mm} 
    \includegraphics[width=\linewidth]{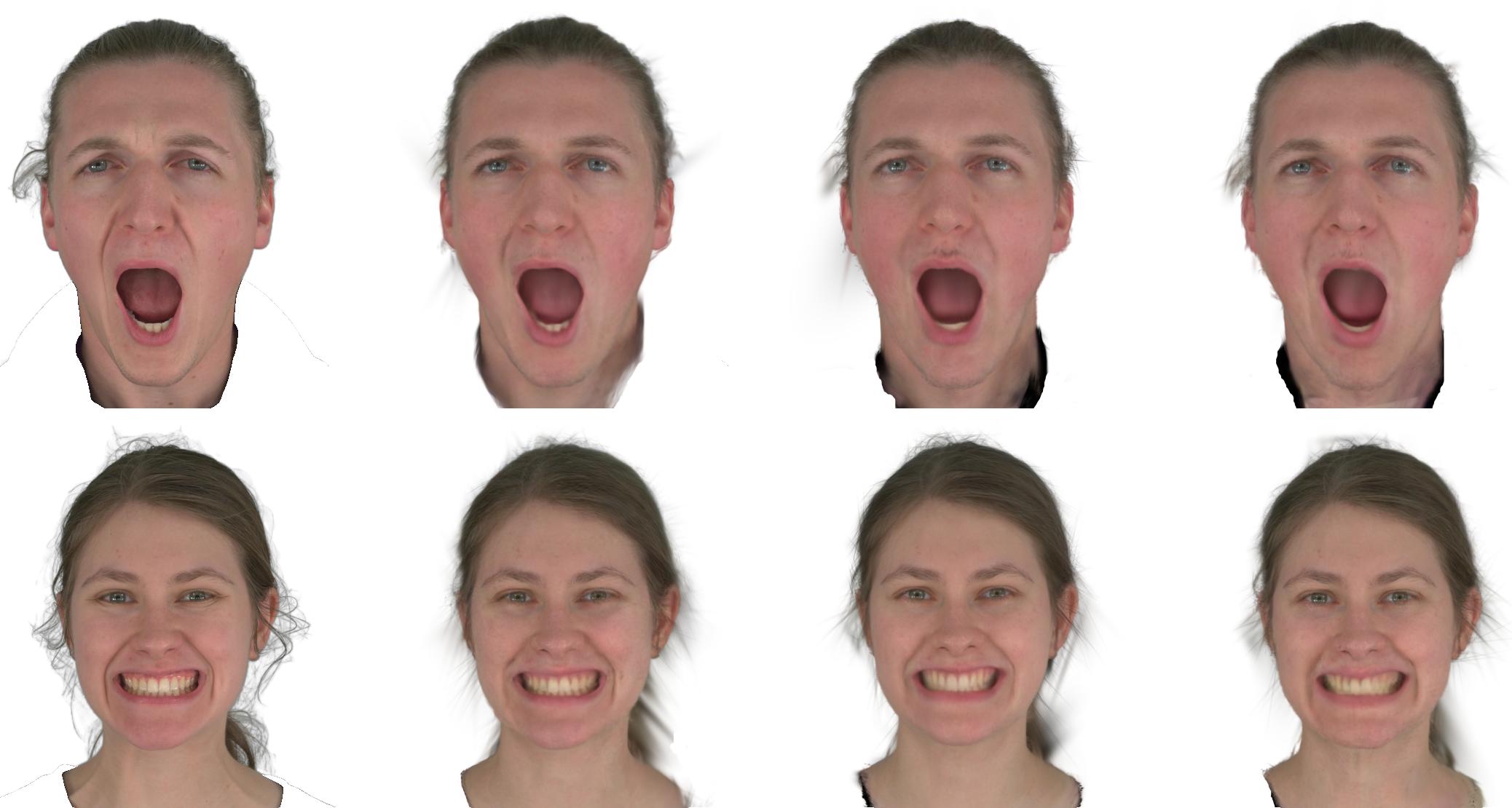}
    \vspace{0.35mm} 
    \begin{tabularx}{\linewidth}{YYYY}
    GT & Ours & DINO & HOG
    \end{tabularx}
    \vspace{-5mm}
    \caption{Comparison of different mouth feature conditionings.}
     \vspace{-5mm}
    \label{fig:DINO}
\end{figure}

\begin{table}[t]
    \caption{Quantitative evaluation of our ablation studies on NeRSemble data.}
    \vspace{-1.2mm}
    \centering
    \small
    \resizebox{\linewidth}{!}{
    \begin{tabular}{|l|c|c|c|c|c|}
    \hline
    \textbf{Method} & \textbf{LPIPS} $\downarrow$ & \textbf{SSIM} $\uparrow$ & \textbf{PSNR} $\uparrow$ & \textbf{ID} $\uparrow$ & \textbf{t-LPIPS} $\downarrow$ \\
    \hline
    \hline
    Ours & 0.064 & 0.937 & 28.016 & 0.991 & 0.079 \\
    Ours-RGB &  0.063 & 0.937 & 28.070 & 0.991 & 0.078 \\
    w/o Adversar. Loss  & 0.070 & 0.939 & 28.137 & 0.990 & 0.084 \\
    w/o Mouth Cond.  & 0.065 & 0.934 & 27.730 & 0.989 & 0.079 \\
    Ours-Full & 0.068 & 0.937 & 27.636 & 0.990 & 0.082 \\
    RGB-Full & 0.061 & 0.941 & 29.632 & 0.991 & 0.074 \\
    \hline
    \end{tabular}}
    \label{tab:ablation}
    \vspace{-1.6mm}
\end{table}
\begin{figure*}[t!]
    \centering
    \vspace{1mm}
    \includegraphics[width=\textwidth]{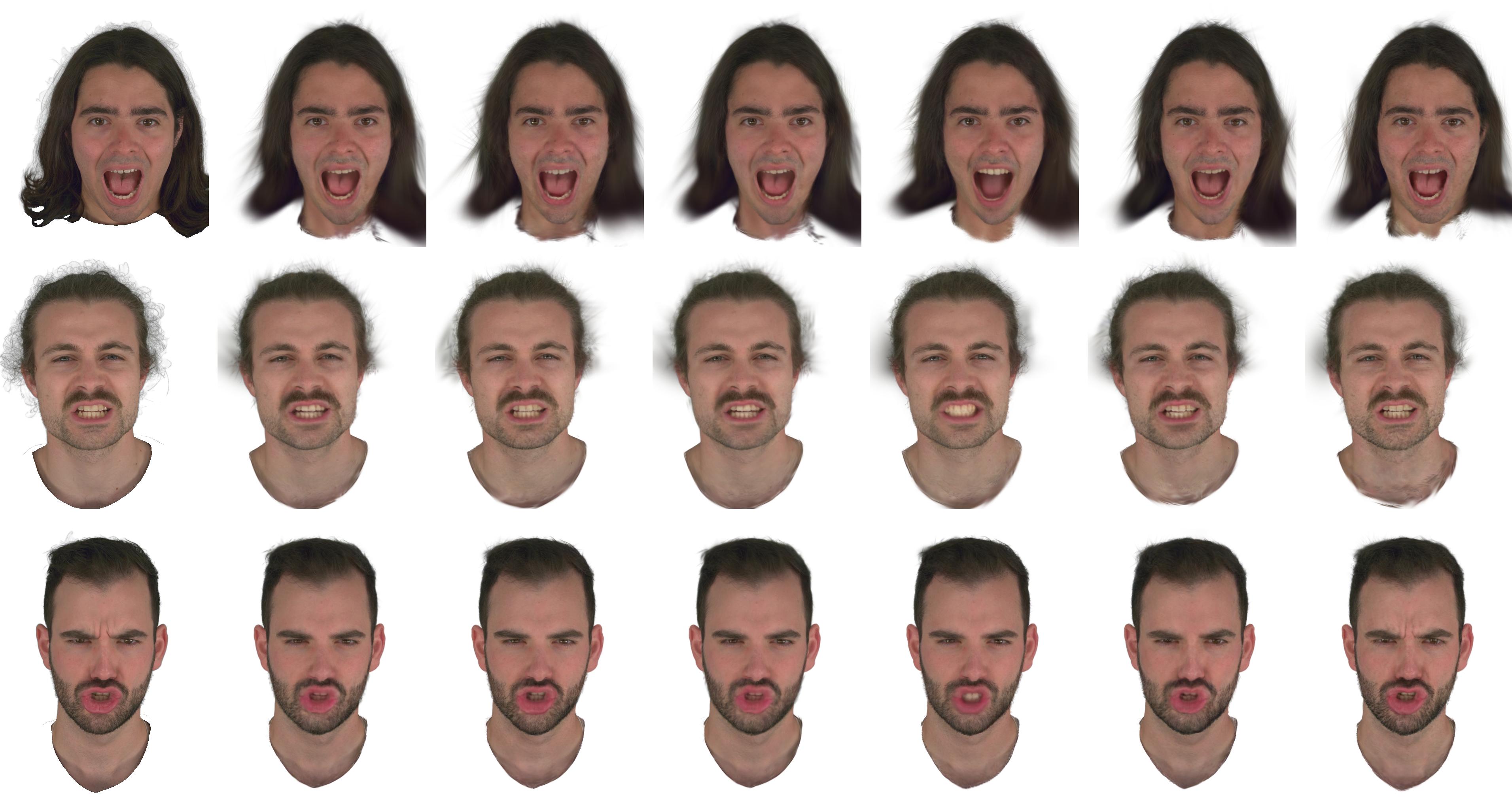}
    \vspace{-1.5mm} 
    \begin{tabularx}{\textwidth}{YYYYYYY}
        Ground Truth & Ours  & Ours-RGB  & w/o Adversarial Loss & w/o Mouth Conditioning & Ours-Full & RGB-Full
    \end{tabularx}
    \caption{Ablation experiments where we replace our gradient-based conditioning with different other types of conditioning, and where we show the influence of the adversarial loss. }
    \label{fig:Ablation}
    \vspace{-2mm}
\end{figure*}
\begin{table}[t]
    \vspace{-1mm}
    \caption{Quantitative evaluation of different mouth feature conditionings on NeRSemble data.}
    \vspace{-1mm}
    \centering
    \small
    \resizebox{\linewidth}{!}{
    \begin{tabular}{|l|c|c|c|c|c|}
    \hline
    \textbf{Method} & \textbf{LPIPS} $\downarrow$ & \textbf{SSIM} $\uparrow$ & \textbf{PSNR} $\uparrow$ & \textbf{ID} $\uparrow$ & \textbf{t-LPIPS} $\downarrow$ \\
    \hline\hline
    Ours & \textbf{0.068}$\pm$0.004 & \textbf{0.932}$\pm$0.002 & \textbf{28.156}$\pm$0.106 & \textbf{0.991}$\pm$0.0006 & \textbf{0.084}$\pm$0.003 \\
    DINO & 0.071$\pm$0.005 & 0.930$\pm$0.001 & 28.037$\pm$0.102 & 0.990$\pm$0.0007 & 0.086$\pm$0.004 \\
    HOG  & 0.079$\pm$0.003 & 0.928$\pm$0.002 & 27.941$\pm$0.109 & 0.990$\pm$0.0003 & 0.089$\pm$0.004 \\
    \hline
    \end{tabular}}
    \label{tab:ablation2}
    \vspace{-3.2mm}
\end{table}
We perform several ablation experiments and evaluate on 20 NeRSemble test subjects with 5 sequences per actor. Firstly, we replace the gradient mouth map with an RGB map. Doing so can copy over the color information from the driving signal thereby limiting the applicability of our technique. We notice from Tab. \ref{tab:ablation} and Fig. \ref{fig:Ablation} (col. 3) that the metrics as well as the visual quality are very similar to our method, though marginally better as expected.
We perform another experiment where we further remove the adversarial loss term during training. The term is important in the sense that it makes our results look perceptually better which is reflected in Tab. \ref{tab:ablation} where our LPIPS loss is lower.
Moreover, in Fig. \ref{fig:Ablation} (col. 4), we can notice that the hair and beard regions look less sharp compared to our method.
Thirdly, we forgo our mouth conditioning and use only SMIRK maps. We can observe in Fig. \ref{fig:Ablation} (col. 5) that the mouth interior does not faithfully model the driving signal and is blurry.
We further conduct two sets of ablations where we condition the entire face region using both gradient and RGB masks instead of just the mouth region. 
We observe from Tab. \ref{tab:ablation} and Fig. \ref{fig:Ablation} (col. 6-7), that the face quality does not increase significantly and is very close to our method even though doing so would undermine the motivation and applicability of our method in a way that one is unable to learn anything from the target identity during inference.
Finally, we conduct an experiment where we use alternative features like DINO \cite{caron2021emerging} and HOG \cite{10.1109/CVPR.2005.177} to condition the mouth region. From Tab. \ref{tab:ablation2} and Fig. \ref{fig:DINO}, we can observe that the gradient-based features are better at reconstructing the mouth region details compared to DINO and HOG.
Thus, we can deduce from these ablation studies that the choice of using gradient-based mouth-only conditioning plus adversarial loss during training, does not impact the quality standards while ensuring the flexibility and applicability of our method.
\subsection{Application}
\begin{figure}[t!]
    \centering
    \includegraphics[width=\linewidth]{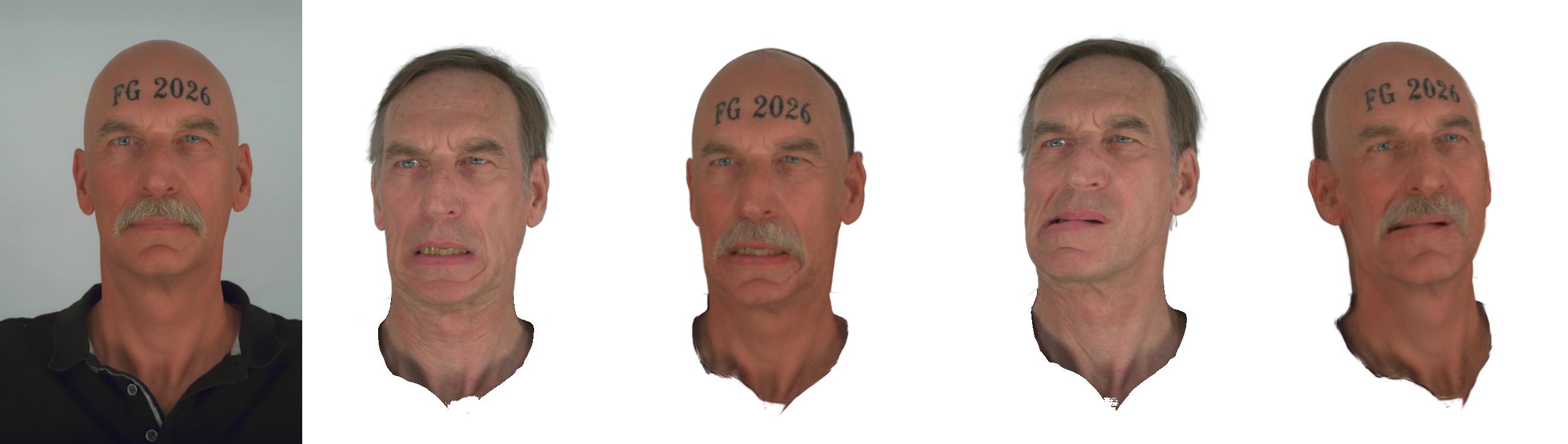}
    \vspace{0.35mm} 
    \begin{tabularx}{\linewidth}{YYYYY}
    Target & Driver  & Result  & Driver & Novel View
    \end{tabularx}
    \label{fig:Application}
    \vspace{-2mm}
    \includegraphics[width=\linewidth]{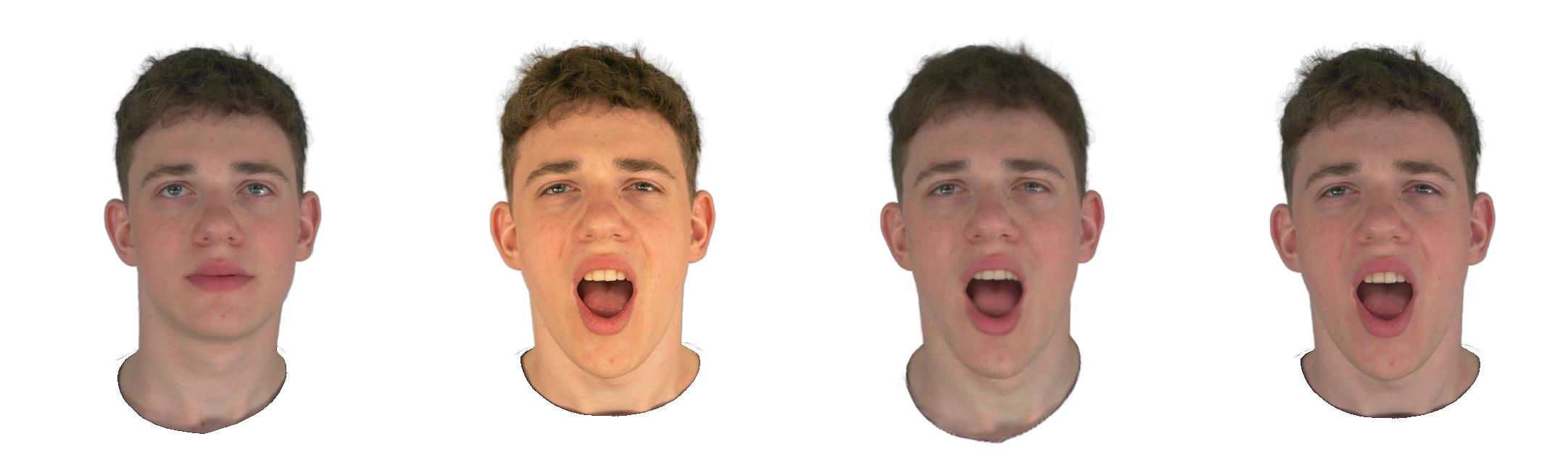}
    \begin{tabularx}{\linewidth}{YYYYYY}
    Target & Driver & Result & GT
    \end{tabularx}
     \vspace{-2mm}
    \caption{Robustness of our method w.r.t. identity and illumination changes. In the first row, the single enrollment image used is an edited image, from which we recover the 3D avatar which can be driven by a monocular sequence. In the bottom row, one can see that the 3D avatar of a person can be driven by a monocular input in different illumination settings.}
    \vspace{-2mm}
    \label{fig:Application2}
\end{figure}

We demonstrate applications of our method by conducting two experiments. Given an image of a target person, we use an online AI-based image editor\footnote{Google AI-Studio} to modify it by adding or removing beard, mustache, tattoo or changing lighting, skin-tone, hair-style, \textit{etc} (see Fig. \ref{fig:Application2} (row 1)). We then perform inversion on this image resulting in a 3D avatar.
We drive the modified avatar using the sequence of the original actor.
We can observe that we achieve high realism with the nuanced expressions being transferred to the avatar.

In a second experiment on a different actor (Fig. \ref{fig:Application2} (row 2)), we obtain an original 3D avatar but simulate changes in illumination in the driving frames. 
We observe that our method is robust to such changes in the driving signal as we are using gradient-based conditioning and thus, the original appearance with a detailed mouth region can be recovered.

\medskip
\noindent \textbf{Limitations.}
Our method is designed for self-reenactment, an application in the scenario of cross-reenactment produces identities that might look realistic but do not reflect the actual appearance of the target identity, especially in the mouth region (note that other methods would tend to produce a generic mouth interior).
The 3DMM-based control of the forehead can lead to missing or reduced wrinkles; depending on the application, one could apply a similar conditioning technique to the forehead region.
Our method cannot handle extreme side-view driving videos and we also noticed some color shift for out-of-distribution data. 
A dataset with more camera view and appearance variety could mitigate this.

\section{CONCLUSIONS}

We have presented 3DRealHead, a novel method for few-shot detailed head avatar reconstruction and animation.
It is based on the inversion of a 3D prior which is trained on a multi-view dataset of human heads (NeRSemble) with adversarial losses.
As a 3D representation, we employ 3D Gaussian primitives which are attached to the 3DMM FLAME that could be efficiently rasterized via 3D Gaussian splatting.
As a driving signal, we not only rely on 3DMM-based facial expressions, but also on appearance features for the mouth region that are directly extracted from the driving video.
The method is designed for the use-case of self-reenactment, however, we also demonstrate cross-reenactment (see Appendix \ref{appendix:Additional Experiments}).
In comparisons, we demonstrate the achieved quality of our method.
We believe that the hybrid of 3DMM-based and feature-based control is the key to achieve believable 3D appearances that could be used for immersive applications in AR and VR.
%


\addtolength{\textheight}{-3cm}


\section{ACKNOWLEDGMENTS}

The authors thank Tobias Kirschstein for running a baseline. The authors gratefully acknowledge the computing time provided on the high-performance computer Lichtenberg at the NHR Center NHR4CES@TUDa. This is funded by the German Federal Ministry of Education and Research (BMBF) and the Hessian Ministry of Science and Research, Art and Culture (HMWK). Justus Thies is supported by the DFG Excellence Strategy— EXC-3057 and the project is co-funded by the European Union (ERC, Lemo, 101162081). Views and opinions expressed are however those of the author(s) only and do not necessarily reflect those of the European Union or the European Research Council. Neither the European Union nor the granting authority can be held responsible for them. The project is also supported by a Google Research Award 2024. All the data were processed outside Google.
%


\section*{ETHICAL IMPACT STATEMENT}

Recovering photo-realistic and animatable representation of humans comes with certain risk of misuse.
State-of-the-art methods target general animation control, demonstrating facial cross-reenactment.
In contrast, our approach focuses on self-reenactment, where facial expression details can be extracted from a 2D video and mapped to a 3D Gaussian primitive-based avatar.
While it also allows for facial cross-reenactment, the resulting identity will not resemble the real subject appearance, as the mouth region will contain features from the driving video.
Still this can result in realistic looking results that can be misused.
To this end, research in forgery / synthetic data detection like~\cite{roessler2019faceforensics++, roessler2018faceforensics} has to be promoted.
%


{\small
\bibliographystyle{ieee}
\bibliography{egbib}

@STRING( CVPR   = "Conference on Computer Vision and Pattern Recognition (CVPR)")

@STRING( CVPRW  = "Conference on Computer Vision and Pattern Recognition Workshops (CVPR-W)")

@STRING( ECCV   = "European Conference on Computer Vision (ECCV)")

@STRING( ICCV   = "International Conference on Computer Vision (ICCV)")

@STRING( I3DV   = "International Conference on 3D Vision (3DV)")

@STRING( NEURIPS = "Advances in Neural Information Processing Systems (NeurIPS)")

@STRING( TOG    = "Transactions on Graphics (TOG)")

@article{thies2019deferred,
  author = {Thies, Justus and Zollh{\"o}fer, Michael and Nie{\ss}ner, Matthias},
  title = {Deferred Neural Rendering: Image Synthesis using Neural Textures},
  journal={ACM Transactions on Graphics 2019 (TOG)},
  year={2019}
}

@Article{kerbl3Dgaussians,
      author       = {Kerbl, Bernhard and Kopanas, Georgios and Leimk{\"u}hler, Thomas and Drettakis, George},
      title        = {3D Gaussian Splatting for Real-Time Radiance Field Rendering},
      journal      = {ACM Transactions on Graphics},
      number       = {4},
      volume       = {42},
      month        = {July},
      year         = {2023},
      url          = {https://repo-sam.inria.fr/fungraph/3d-gaussian-splatting/}
}

@inproceedings{Zielonka2022InstantVH,
  title     = {Instant Volumetric Head Avatars},
  author    = {Wojciech Zielonka and Timo Bolkart and Justus Thies},
  booktitle = {CVPR},
  year      = {2023},
  pages     = {4574-4584},
}

@inproceedings{xiang2024flashavatar,
      author    = {Jun Xiang and Xuan Gao and Yudong Guo and Juyong Zhang},
      title     = {FlashAvatar: High-fidelity Head Avatar with Efficient Gaussian Embedding},
      booktitle = {The IEEE Conference on Computer Vision and Pattern Recognition (CVPR)},
      year      = {2024},
  }

@inproceedings{shao2024splattingavatar,
  title = {{SplattingAvatar: Realistic Real-Time Human Avatars with Mesh-Embedded Gaussian Splatting}},
  author = {Shao, Zhijing and Wang, Zhaolong and Li, Zhuang and Wang, Duotun and Lin, Xiangru and Zhang, Yu and Fan, Mingming and Wang, Zeyu},
  booktitle = {Proceedings of the IEEE/CVF Conference on Computer Vision and Pattern Recognition (CVPR)},
  year = {2024}
}

@misc{an2023panohead,
      title={PanoHead: Geometry-Aware 3D Full-Head Synthesis in 360$^{\circ}$}, 
      author={Sizhe An and Hongyi Xu and Yichun Shi and Guoxian Song and Umit Ogras and Linjie Luo},
      year={2023},
      eprint={2303.13071},
      archivePrefix={arXiv},
      primaryClass={cs.CV}
}

@inproceedings{zielonka2024gem,
    title     = {Gaussian Eigen Models for Human Heads},
    author    = {Wojciech Zielonka and Timo Bolkart and Thabo Beeler and Justus Thies},
    booktitle = {CVPR},
    month     = {June},
    year      = {2025},
}

@inproceedings{xu2023gaussianheadavatar,
  title={Gaussian Head Avatar: Ultra High-fidelity Head Avatar via Dynamic Gaussians},
  author={Xu, Yuelang and Chen, Benwang and Li, Zhe and Zhang, Hongwen and Wang, Lizhen and Zheng, Zerong and Liu, Yebin},
  booktitle={Proceedings of the IEEE/CVF Conference on Computer Vision and Pattern Recognition (CVPR)},
  year={2024}
}

@inproceedings{qian2024gaussianavatars,
  title={Gaussianavatars: Photorealistic head avatars with rigged 3d gaussians},
  author={Qian, Shenhan and Kirschstein, Tobias and Schoneveld, Liam and Davoli, Davide and Giebenhain, Simon and Nie{\ss}ner, Matthias},
  booktitle={Proceedings of the IEEE/CVF Conference on Computer Vision and Pattern Recognition},
  pages={20299--20309},
  year={2024}
}

@misc{zheng2025headgapfewshot3dhead,
      title={HeadGAP: Few-Shot 3D Head Avatar via Generalizable Gaussian Priors}, 
      author={Xiaozheng Zheng and Chao Wen and Zhaohu Li and Weiyi Zhang and Zhuo Su and Xu Chang and Yang Zhao and Zheng Lv and Xiaoyuan Zhang and Yongjie Zhang and Guidong Wang and Lan Xu},
      year={2025},
      eprint={2408.06019},
      archivePrefix={arXiv},
      primaryClass={cs.CV},
      url={https://arxiv.org/abs/2408.06019}, 
}

@inproceedings{saito2024rgca,
          author = {Shunsuke Saito and Gabriel Schwartz and Tomas Simon and Junxuan Li and Giljoo Nam},
          title = {Relightable Gaussian Codec Avatars}, 
          booktitle = {CVPR},
          year = {2024},
        }

@inproceedings{kirschstein2024diffusionavatars,
  title={Diffusionavatars: Deferred diffusion for high-fidelity 3d head avatars},
  author={Kirschstein, Tobias and Giebenhain, Simon and Nie{\ss}ner, Matthias},
  booktitle={Proceedings of the IEEE/CVF Conference on Computer Vision and Pattern Recognition},
  pages={5481--5492},
  year={2024}
}

@inproceedings{wang2023styleavatar,
  title={StyleAvatar: Real-time Photo-realistic Portrait Avatar from a Single Video},
  author={Wang, Lizhen and Zhao, Xiaochen and Sun, Jingxiang and Zhang, Yuxiang and Zhang, Hongwen and Yu, Tao and Liu, Yebin},
  booktitle={ACM SIGGRAPH 2023 Conference Proceedings},
  pages={},
  year={2023}
}

@inproceedings{kabadayi24ganavatar,
      title = {GAN-Avatar: Controllable Personalized GAN-based Human Head Avatar},
      author = {Kabadayi, Berna and Zielonka, Wojciech and Bhatnagar, Bharat Lal  and Pons-Moll, Gerard and Thies, Justus},
      booktitle = {International Conference on 3D Vision (3DV)},
      month = {March},
      year = {2024},
  }

@inproceedings{zielonka25dega,
  title        = {Drivable 3D Gaussian Avatars},
  author       = {Wojciech Zielonka and Timur Bagautdinov and Shunsuke Saito and Michael Zollhöfer and Justus Thies and Javier Romero},
  booktitle    = {I3DV},
  month        = {March},
  year         = {2025}
}

@inproceedings{kirschstein2024gghead,
    author = {Kirschstein, Tobias and Giebenhain, Simon and Tang, Jiapeng and Georgopoulos, Markos and Nie\ss{}ner, Matthias},
    title = {{GGHead: Fast and Generalizable 3D Gaussian Heads}},
    year = {2024},
    isbn = {9798400711312},
    publisher = {Association for Computing Machinery},
    address = {New York, NY, USA},
    url = {https://doi.org/10.1145/3680528.3687686},
    doi = {10.1145/3680528.3687686},
    booktitle = {SIGGRAPH Asia 2024 Conference Papers},
    articleno = {126},
    numpages = {11},
    series = {SA '24}
}

@incollection{buehler2024cafca,
    title={Cafca: High-quality Novel View Synthesis of Expressive Faces from Casual Few-shot Captures},
    author={Marcel C. Buehler and Gengyan Li and Erroll Wood and Leonhard Helminger and Xu Chen and Tanmay Shah and Daoye Wang and Stephan Garbin and Sergio Orts-Escolano and Otmar Hilliges and Dmitry Lagun and Jérémy Riviere and Paulo Gotardo and Thabo Beeler and Abhimitra Meka and Kripasindhu Sarkar},
    year={2024},
    booktitle={ACM SIGGRAPH Asia 2024 Conference Paper},
    doi={10.1145/3680528.3687580},
    url={https://doi.org/10.1145/3680528}
}

@inproceedings{buhler2023preface,
  title={Preface: A Data-driven Volumetric Prior for Few-shot Ultra High-resolution Face Synthesis},
  author={B{\"u}hler, Marcel C and Sarkar, Kripasindhu and Shah, Tanmay and Li, Gengyan and Wang, Daoye and Helminger, Leonhard and Orts-Escolano, Sergio and Lagun, Dmitry and Hilliges, Otmar and Beeler, Thabo and others},
  booktitle={Proceedings of the IEEE/CVF International Conference on Computer Vision},
  pages={3402--3413},
  year={2023}
}

@inproceedings{
    chu2024gagavatar,
    title={Generalizable and Animatable Gaussian Head Avatar},
    author={Xuangeng Chu and Tatsuya Harada},
    booktitle=NEURIPS,
    year={2024},
}

@misc{gafni2020dynamicneuralradiancefields,
      title={Dynamic Neural Radiance Fields for Monocular 4D Facial Avatar Reconstruction}, 
      author={Guy Gafni and Justus Thies and Michael Zollhöfer and Matthias Nießner},
      year={2020},
      eprint={2012.03065},
      archivePrefix={arXiv},
      primaryClass={cs.CV},
      url={https://arxiv.org/abs/2012.03065}, 
}

@inproceedings{thies2016face,
   title       = {{Face2Face: Real-time Face Capture and Reenactment of RGB Videos}},
   author      = {Thies, J. and Zollh{\"o}fer, M. and Stamminger, M. and Theobalt, C. and Nie{\ss}ner, M.},
   booktitle   = {Proc. Computer Vision and Pattern Recognition (CVPR), IEEE},
   year        = {2016}
}

@article{Li2017flame, 
  title = {Learning a model of facial shape and expression from {4D} scans}, 
  author = {Li, Tianye and Bolkart, Timo and Black, Michael. J. and Li, Hao and Romero, Javier}, 
  journal = {ACM Transactions on Graphics, (Proc. SIGGRAPH Asia)}, 
  volume = {36}, 
  number = {6}, 
  year = {2017}, 
  pages = {194:1--194:17},
  url = {https://doi.org/10.1145/3130800.3130813} 
}

@misc{johnson2016perceptuallossesrealtimestyle,
      title={Perceptual Losses for Real-Time Style Transfer and Super-Resolution}, 
      author={Justin Johnson and Alexandre Alahi and Li Fei-Fei},
      year={2016},
      eprint={1603.08155},
      archivePrefix={arXiv},
      primaryClass={cs.CV},
      url={https://arxiv.org/abs/1603.08155}, 
}

@inproceedings{mildenhall2020nerf,
  title={NeRF: Representing Scenes as Neural Radiance Fields for View Synthesis},
  author={Ben Mildenhall and Pratul P. Srinivasan and Matthew Tancik and Jonathan T. Barron and Ravi Ramamoorthi and Ren Ng},
  year={2020},
  booktitle={ECCV},
}

@article{Chen2023MonoGaussianAvatarMG,
  title={MonoGaussianAvatar: Monocular Gaussian Point-based Head Avatar},
  author={Yufan Chen and Lizhen Wang and Qijing Li and Hongjiang Xiao and Shengping Zhang and Hongxun Yao and Yebin Liu},
  journal={ACM SIGGRAPH 2024 Conference Papers},
  year={2023},
  url={https://api.semanticscholar.org/CorpusID:266053803}
}

@misc{chu2024gpavatargeneralizableprecisehead,
      title={GPAvatar: Generalizable and Precise Head Avatar from Image(s)}, 
      author={Xuangeng Chu and Yu Li and Ailing Zeng and Tianyu Yang and Lijian Lin and Yunfei Liu and Tatsuya Harada},
      year={2024},
      eprint={2401.10215},
      archivePrefix={arXiv},
      primaryClass={cs.CV},
      url={https://arxiv.org/abs/2401.10215}, 
}

@misc{yu2024one2avatargenerativeimplicithead,
      title={One2Avatar: Generative Implicit Head Avatar For Few-shot User Adaptation}, 
      author={Zhixuan Yu and Ziqian Bai and Abhimitra Meka and Feitong Tan and Qiangeng Xu and Rohit Pandey and Sean Fanello and Hyun Soo Park and Yinda Zhang},
      year={2024},
      eprint={2402.11909},
      archivePrefix={arXiv},
      primaryClass={cs.CV},
      url={https://arxiv.org/abs/2402.11909}, 
}

@article{chen2024morphable,
      title={Morphable Diffusion: 3D-Consistent Diffusion for Single-image Avatar Creation}, 
      author={Xiyi Chen and Marko Mihajlovic and Shaofei Wang and Sergey Prokudin and Siyu Tang},
      booktitle={IEEE Conference on Computer Vision and Pattern Recognition (CVPR)},
      year={2024}
      }

@inproceedings{zhang2025rodinhd,
  title={Rodinhd: High-fidelity 3d avatar generation with diffusion models},
  author={Zhang, Bowen and Cheng, Yiji and Wang, Chunyu and Zhang, Ting and Yang, Jiaolong and Tang, Yansong and Zhao, Feng and Chen, Dong and Guo, Baining},
  booktitle={European Conference on Computer Vision},
  pages={465--483},
  year={2025},
  organization={Springer}
}

@inproceedings{wang2023rodin,
author = {Wang, Tengfei and Zhang, Bo and Zhang, Ting and Gu, Shuyang and Bao, Jianmin and Baltrusaitis, Tadas and Shen, Jingjing and Chen, Dong and Wen, Fang and Chen, Qifeng and Guo, Baining},
year = {2023},
month = {06},
pages = {4563-4573},
title = {RODIN: A Generative Model for Sculpting 3D Digital Avatars Using Diffusion},
doi = {10.1109/CVPR52729.2023.00443}
}

@inproceedings{Chan2021,
author = {Chan, Eric and Lin, Connor and Chan, Matthew and Nagano, Koki and Pan, Boxiao and Mello, Shalini and Gallo, Orazio and Guibas, Leonidas and Tremblay, Jonathan and Khamis, Sameh and Karras, Tero and Wetzstein, Gordon},
year = {2022},
month = {06},
pages = {16102-16112},
title = {Efficient Geometry-aware 3D Generative Adversarial Networks},
doi = {10.1109/CVPR52688.2022.01565}
}

@inproceedings{lan2023gaussian3diff,
  author={Lan, Yushi and Tan, Feitong and Qiu, Di and Xu, Qiangeng and Genova, Kyle and Huang, Zeng and Fanello, Sean and Pandey, Rohit and Funkhouser, Thomas and Loy, Chen Change and Zhang, Yinda},
  title={Gaussian3Diff: 3D Gaussian Diffusion for 3D Full Head Synthesis and Editing},
  year={2024},
  booktitle={ECCV},
}

@inproceedings{10.1145/311535.311556,
author = {Blanz, Volker and Vetter, Thomas},
title = {A morphable model for the synthesis of 3D faces},
year = {1999},
isbn = {0201485605},
publisher = {ACM Press/Addison-Wesley Publishing Co.},
address = {USA},
url = {https://doi.org/10.1145/311535.311556},
doi = {10.1145/311535.311556},
booktitle = {Proceedings of the 26th Annual Conference on Computer Graphics and Interactive Techniques},
pages = {187–194},
numpages = {8},
series = {SIGGRAPH '99}
}

@misc{egger20203dmorphablefacemodels,
      title={3D Morphable Face Models -- Past, Present and Future}, 
      author={Bernhard Egger and William A. P. Smith and Ayush Tewari and Stefanie Wuhrer and Michael Zollhoefer and Thabo Beeler and Florian Bernard and Timo Bolkart and Adam Kortylewski and Sami Romdhani and Christian Theobalt and Volker Blanz and Thomas Vetter},
      year={2020},
      eprint={1909.01815},
      archivePrefix={arXiv},
      primaryClass={cs.CV},
      url={https://arxiv.org/abs/1909.01815}, 
}

@inproceedings{roessler2019faceforensics++,
	author = {Andreas R\"ossler and Davide Cozzolino and Luisa Verdoliva and Christian Riess and Justus Thies and Matthias Nie{\ss}ner},
	title = {FaceForensics++: Learning to Detect Manipulated Facial Images},
	booktitle = {ICCV 2019},
	year={2019}
}

@article{roessler2018faceforensics,
	author = {Andreas R\"ossler and Davide Cozzolino and Luisa Verdoliva and Christian Riess and Justus Thies and Matthias Nie{\ss}ner},
	title = {FaceForensics: A Large-scale Video Dataset for Forgery Detection in Human Faces},
	journal={arXiv},
	year={2018}
}

@article{Lombardi21,
 author = {Lombardi, Stephen and Simon, Tomas and Schwartz, Gabriel and Zollhoefer, Michael and Sheikh, Yaser and Saragih, Jason},
 title = {Mixture of Volumetric Primitives for Efficient Neural Rendering},
 year = {2021},
 issue_date = {August 2021},
 publisher = {Association for Computing Machinery},
 address = {New York, NY, USA},
 volume = {40},
 number = {4},
 issn = {0730-0301},
 url = {https://doi.org/10.1145/3450626.3459863},
 doi = {10.1145/3450626.3459863},
 journal = {ACM Trans. Graph.},
 month = {jul},
 articleno = {59},
 numpages = {13}
}

@article{teotia24gaussianheads,
  title={GaussianHeads: End-to-End Learning of Drivable Gaussian Head Avatars from Coarse-to-fine Representations},
  author={Kartik Teotia and Hyeongwoo Kim and Pablo Garrido and Marc Habermann and Mohamed Elgharib and Christian Theobalt},
  journal={ACM Transactions on Graphics (TOG)},
  year={2024},
  volume={43},
  pages={1 - 12},
  url={https://api.semanticscholar.org/CorpusID:272707579}
}

@article{Kingma2014AdamAM,
  title={Adam: A Method for Stochastic Optimization},
  author={Diederik P. Kingma and Jimmy Ba},
  journal={CoRR},
  year={2014},
  volume={abs/1412.6980},
  url={https://api.semanticscholar.org/CorpusID:6628106}
}

@misc{deng2024portrait4dlearningoneshot4d,
      title={Portrait4D: Learning One-Shot 4D Head Avatar Synthesis using Synthetic Data}, 
      author={Yu Deng and Duomin Wang and Xiaohang Ren and Xingyu Chen and Baoyuan Wang},
      year={2024},
      eprint={2311.18729},
      archivePrefix={arXiv},
      primaryClass={cs.CV},
      url={https://arxiv.org/abs/2311.18729}, 
}

@inproceedings{invertavatar,
author = {Zhao, Xiaochen and Sun, Jingxiang and Wang, Lizhen and Suo, Jinli and Liu, Yebin},
title = {InvertAvatar: Incremental GAN Inversion for Generalized Head Avatars},
year = {2024},
isbn = {9798400705250},
publisher = {Association for Computing Machinery},
address = {New York, NY, USA},
url = {https://doi.org/10.1145/3641519.3657478},
doi = {10.1145/3641519.3657478},
booktitle = {ACM SIGGRAPH 2024 Conference Papers},
articleno = {59},
numpages = {10},
location = {Denver, CO, USA},
series = {SIGGRAPH '24}
}

@inproceedings{sun2023next3d,
      author = {Sun, Jingxiang and Wang, Xuan and Wang, Lizhen and Li, Xiaoyu and Zhang, Yong and Zhang, Hongwen and Liu, Yebin},
      title = {Next3D: Generative Neural Texture Rasterization for 3D-Aware Head Avatars},
      booktitle = {CVPR},
      year = {2023}
}

@article{kirschstein2023nersemble,
    author = {Kirschstein, Tobias and Qian, Shenhan and Giebenhain, Simon and Walter, Tim and Nie\ss{}ner, Matthias},
    title = {NeRSemble: Multi-View Radiance Field Reconstruction of Human Heads},
    year = {2023},
    issue_date = {August 2023},
    publisher = {Association for Computing Machinery},
    address = {New York, NY, USA},
    volume = {42},
    number = {4},
    issn = {0730-0301},
    url = {https://doi.org/10.1145/3592455},
    doi = {10.1145/3592455},
    journal = {ACM Trans. Graph.},
    month = {jul},
    articleno = {161},
    numpages = {14},
}

@inproceedings{zielonka2025synshot,
  title     = {Synthetic Prior for Few-Shot Drivable Head Avatar Inversion},
  author    = {Wojciech Zielonka and Stephan J. Garbin and Alexandros Lattas and George Kopanas and Paulo Gotardo and Thabo Beeler and Justus Thies and Timo Bolkart},
  booktitle = {CVPR},
  month     = {June},
  year      = {2025},
}

@inproceedings{SMIRK:CVPR:2024,
    title = {3D Facial Expressions through Analysis-by-Neural-Synthesis},
    author = {Retsinas, George and Filntisis, Panagiotis P. and Danecek, Radek and Abrevaya, Victoria F. and Roussos, Anastasios and Bolkart, Timo and Maragos, Petros},
    booktitle = {Conference on Computer Vision and Pattern Recognition (CVPR)},
    year = {2024}
}

@inproceedings{caron2021emerging,
  title={Emerging Properties in Self-Supervised Vision Transformers},
  author={Caron, Mathilde and Touvron, Hugo and Misra, Ishan and J\'egou, Herv\'e  and Mairal, Julien and Bojanowski, Piotr and Joulin, Armand},
  booktitle={Proceedings of the International Conference on Computer Vision (ICCV)},
  year={2021}
}

@InProceedings{Sauer2023ICML,
  author    = {Axel Sauer and Tero Karras and Samuli Laine and Andreas Geiger and Timo Aila},
  title     = {{StyleGAN-T}: Unlocking the Power of {GANs} for Fast Large-Scale Text-to-Image Synthesis},
  journal   = {International Conference on Machine Learning},
  volume    = {abs/2301.09515},
  year      = {2023},
  url       = {https://arxiv.org/abs/2301.09515},
}

@misc{qian2024vhap,
  title={VHAP: Versatile Head Alignment with Adaptive Appearance Priors},
  author={Qian, Shenhan},
  year={2024},
  month={sep},
  doi={10.5281/zenodo.14988309},
  url={https://github.com/ShenhanQian/VHAP}
}

@misc{goodfellow2014generativeadversarialnetworks,
      title={Generative Adversarial Networks}, 
      author={Ian J. Goodfellow and Jean Pouget-Abadie and Mehdi Mirza and Bing Xu and David Warde-Farley and Sherjil Ozair and Aaron Courville and Yoshua Bengio},
      year={2014},
      eprint={1406.2661},
      archivePrefix={arXiv},
      primaryClass={stat.ML},
      url={https://arxiv.org/abs/1406.2661}, 
}

@article{wang2004image,
author = {Wang, Zhou and Bovik, Alan and Sheikh, Hamid and Simoncelli, Eero},
year = {2004},
month = {05},
pages = {600 - 612},
title = {Image Quality Assessment: From Error Visibility to Structural Similarity},
volume = {13},
journal = {Image Processing, IEEE Transactions on},
doi = {10.1109/TIP.2003.819861}
}

@article{deep_appearance_models,
 author = {Lombardi, Stephen and Saragih, Jason and Simon, Tomas and Sheikh, Yaser},
 title = {Deep Appearance Models for Face Rendering},
 journal = {ACM Trans. Graph.},
 issue_date = {August 2018},
 volume = {37},
 number = {4},
 month = jul,
 year = {2018},
 issn = {0730-0301},
 pages = {68:1--68:13},
 articleno = {68},
 numpages = {13},
 publisher = {ACM},
 address = {New York, NY, USA},
}

@inproceedings{latentAvatar,
author = {Xu, Yuelang and Zhang, Hongwen and Wang, Lizhen and Zhao, Xiaochen and Huang, Han and Qi, Guojun and Liu, Yebin},
title = {LatentAvatar: Learning Latent Expression Code for Expressive Neural Head Avatar},
year = {2023},
isbn = {9798400701597},
publisher = {Association for Computing Machinery},
address = {New York, NY, USA},
url = {https://doi.org/10.1145/3588432.3591545},
doi = {10.1145/3588432.3591545},
booktitle = {ACM SIGGRAPH 2023 Conference Proceedings},
articleno = {86},
numpages = {10},
location = {Los Angeles, CA, USA},
series = {SIGGRAPH '23}
}

@inproceedings{NEURIPS2024_9712b783,
 author = {Martinez, Julieta and Kim, Emily and Romero, Javier and Bagautdinov, Timur and Saito, Shunsuke and Yu, Shoou-I and Anderson, Stuart and Zollh\"{o}fer, Michael and Wang, Te-Li and Bai, Shaojie and Li, Chenghui and Wei, Shih-En and Joshi, Rohan and Borsos, Wyatt and Simon, Tomas and Saragih, Jason and Theodosis, Paul and Greene, Alexander and Josyula, Anjani and Maeta, Silvio Mano and Jewett, Andrew I. and Venshtain, Simon and Heilman, Christopher and Chen, Yueh-Tung and Fu, Sidi and Elshaer, Mohamed Ezzeldin A. and Du, Tingfang and Wu, Longhua and Chen, Shen-Chi and Kang, Kai and Wu, Michael and Emad, Youssef and Longay, Steven and Brewer, Ashley and Shah, Hitesh and Booth, James and Koska, Taylor and Haidle, Kayla and Andromalos, Matt and Hsu, Joanna and Dauer, Thomas and Selednik, Peter and Godisart, Tim and Ardisson, Scott and Cipperly, Matthew and Humberston, Ben and Farr, Lon and Hansen, Bob and Guo, Peihong and Braun, Dave and Krenn, Steven and Wen, He and Evans, Lucas and Fadeeva, Natalia and Stewart, Matthew and Schwartz, Gabriel and Gupta, Divam and Moon, Gyeongsik and Guo, Kaiwen and Dong, Yuan and Xu, Yichen and Shiratori, Takaaki and Prada, Fabian and Pires, Bernardo R. and Peng, Bo and Buffalini, Julia and Trimble, Autumn and McPhail, Kevyn and Schoeller, Melissa and Sheikh, Yaser},
 booktitle = {Advances in Neural Information Processing Systems},
 doi = {10.52202/079017-2640},
 editor = {A. Globerson and L. Mackey and D. Belgrave and A. Fan and U. Paquet and J. Tomczak and C. Zhang},
 pages = {83008--83023},
 publisher = {Curran Associates, Inc.},
 title = {Codec Avatar Studio: Paired Human Captures for Complete, Driveable, and Generalizable Avatars},
 url = {https://proceedings.neurips.cc/paper_files/paper/2024/file/9712b78386cebdc3db7f1a48c2d20edb-Paper-Datasets_and_Benchmarks_Track.pdf},
 volume = {37},
 year = {2024}
}

@inproceedings{wuu2022multiface,
  title={Multiface: A Dataset for Neural Face Rendering},
  author = {Wuu, Cheng-hsin and Zheng, Ningyuan and Ardisson, Scott and Bali, Rohan and Belko, Danielle and Brockmeyer, Eric and Evans, Lucas and Godisart, Timothy and Ha, Hyowon and Huang, Xuhua and Hypes, Alexander and Koska, Taylor and Krenn, Steven and Lombardi, Stephen and Luo, Xiaomin and McPhail, Kevyn and Millerschoen, Laura and Perdoch, Michal and Pitts, Mark and Richard, Alexander and Saragih, Jason and Saragih, Junko and Shiratori, Takaaki and Simon, Tomas and Stewart, Matt and Trimble, Autumn and Weng, Xinshuo and Whitewolf, David and Wu, Chenglei and Yu, Shoou-I and Sheikh, Yaser},
  booktitle={arXiv},
  year={2022},
  doi = {10.48550/ARXIV.2207.11243},
  url = {https://arxiv.org/abs/2207.11243}
}

@article{mueller2022instant,
    author = {Thomas M\"uller and Alex Evans and Christoph Schied and Alexander Keller},
    title = {Instant Neural Graphics Primitives with a Multiresolution Hash Encoding},
    journal = {ACM Trans. Graph.},
    issue_date = {July 2022},
    volume = {41},
    number = {4},
    month = jul,
    year = {2022},
    pages = {102:1--102:15},
    articleno = {102},
    numpages = {15},
    url = {https://doi.org/10.1145/3528223.3530127},
    doi = {10.1145/3528223.3530127},
    publisher = {ACM},
    address = {New York, NY, USA}
}

@misc{he2025lamlargeavatarmodel,
      title={LAM: Large Avatar Model for One-shot Animatable Gaussian Head}, 
      author={Yisheng He and Xiaodong Gu and Xiaodan Ye and Chao Xu and Zhengyi Zhao and Yuan Dong and Weihao Yuan and Zilong Dong and Liefeng Bo},
      year={2025},
      eprint={2502.17796},
      archivePrefix={arXiv},
      primaryClass={cs.CV},
      url={https://arxiv.org/abs/2502.17796}, 
}

@InProceedings{xie2022vfhq,
      author = {Liangbin Xie and Xintao Wang and Honglun Zhang and Chao Dong and Ying Shan},
      title = {VFHQ: A High-Quality Dataset and Benchmark for Video Face Super-Resolution},
      booktitle={The IEEE Conference on Computer Vision and Pattern Recognition Workshops (CVPRW)},
      year = {2022}
  }

@inproceedings{10.1145/2659467.2675048,
author = {Jacobson, Alec and Gingold, Yotam},
title = {Skinning: real-time shape deformation},
year = {2014},
isbn = {9781450331951},
publisher = {Association for Computing Machinery},
address = {New York, NY, USA},
url = {https://doi.org/10.1145/2659467.2675048},
doi = {10.1145/2659467.2675048},
booktitle = {SIGGRAPH Asia 2014 Courses},
articleno = {19},
location = {Shenzhen, China},
series = {SA '14}
}

@InProceedings{kirschstein2025avat3r,
    author    = {Kirschstein, Tobias and Romero, Javier and Sevastopolsky, Artem and Nie{\ss}ner, Matthias and Saito, Shunsuke},
    title     = {Avat3r: Large Animatable Gaussian Reconstruction Model for High-fidelity 3D Head Avatars},
    booktitle = {Proceedings of the IEEE/CVF International Conference on Computer Vision (ICCV)},
    month     = {October},
    year      = {2025},
    pages     = {12089-12100}
}

@inproceedings{Sobel1990AnI3,
  title={An Isotropic 3×3 image gradient operator},
  author={Irwin Sobel and G. M. Feldman},
  year={1990},
  url={https://api.semanticscholar.org/CorpusID:59909525}
}

@inproceedings{10.1109/CVPR.2005.177,
author = {Dalal, Navneet and Triggs, Bill},
title = {Histograms of Oriented Gradients for Human Detection},
year = {2005},
isbn = {0769523722},
publisher = {IEEE Computer Society},
address = {USA},
url = {https://doi.org/10.1109/CVPR.2005.177},
doi = {10.1109/CVPR.2005.177},
booktitle = {Proceedings of the 2005 IEEE Computer Society Conference on Computer Vision and Pattern Recognition (CVPR'05) - Volume 1 - Volume 01},
pages = {886–893},
numpages = {8},
series = {CVPR '05}
}
}

\newpage
\title{\LARGE \bf
3DRealHead: Few-Shot Detailed Head Avatar
}
\author{\Large Supplemental Document}

\maketitle
\thispagestyle{plain}
\pagestyle{plain}

\def\thesection{\Alph{section}}
\setcounter{section}{0}
\section{Architecture Details}
Our model employs a Style U-Net \cite{wang2023styleavatar} with two identical encoders and a common decoder. The two encoders take as input identity and expression conditioning signals, respectively, in the form of texture and position maps defined in a given UV space.
The encoder is a downsampling network with 6 convolutional blocks that generate a hierarchy of feature maps ranging from $8 \times 8$ to $256 \times 256$ spatial dimensions. The channel size decreases with an increase in spatial dimension. The feature maps are passed as skip connections into the decoder. 
The decoder consists of a series of upsampling convolutional blocks. It takes as input the final encoder feature map and passes it through the upsampling layers.
It also concatenates the feature maps of the corresponding encoder output layer in the form of skip connections to the output of the previous upsampling layer before sending them to the next block.
Since our model comprises two encoders, we additionally concatenate the identity and expression skip feature maps together before passing them into the next upsampling block in the decoder.
A final layer of convolution is applied to the last layer to generate the output map with a spatial resolution of $512 \times 512$.
Each pixel has a channel size of 59 which is equal to the number of parameters of each 3D Gaussian.

\section{Inference Details}
\label{appendix:Inference}
During the enrollment stage, given the target actor image, we optimize the identity encoder output, \textit{i.e.}, the hierarchical feature maps along with the last layer of the decoder network.
The remaining part of the model is frozen, thus preserving the prior information.
We perform fine-tuning in a two stage approach.
In the first stage, we only optimize for the lower 3 out of the 6 feature maps. In the second stage, we optimize for the remaining 3 high-level feature maps and the final decoder layer.
We observe that fine-tuning in such a two-stage manner results in a quality similar to that of fine-tuning the entire decoder.
Additionally, for out-of-distribution data, we also optimize for an affine transform matrix $C_{col} \in \mathbb{R}^{4x4}$ representing the color-shift between the rendered and ground truth images.

\section{Image Gradient Maps}
The image gradient maps are provided as additional inputs to the expression encoder for extracting information about the mouth region. The 3-channel gradient maps are obtained from RGB driving frames by first computing the image gradients. We do so by applying Sobel filters \cite{Sobel1990AnI3} to the image in the horizontal (x-axis) and vertical (y-axis) directions. The RGB image is first converted to grayscale before applying the filter. The gradients are scaled accordingly in the [-1,1] range by dividing with the maximum absolute value. In addition, the magnitude of the computed gradients is also stored in the third channel. Finally, using the tracked meshes, the image gradient is stored in the UV space.
\section{Additional Experiments}
\label{appendix:Additional Experiments}
In this section, we perform several experiments where we demonstrate the smoothness of our latent identity features, show cross-reenactment results, perform an experiment showing the effect of fine-tuning during inversion, show an additional comparison and show some view-synthesis results on In-the-Wild data. We further perform experiments analyzing the distribution of Gaussians in the mouth region, the impact of using additional mouth conditioning, the size of the training set, category-wise performance, the number of frames used during enrollment and an additional comparison against GaussianAvatars \cite{qian2024gaussianavatars}.
\\ \\
\subsection{Identity Interpolation}
\begin{figure}[t!]
    \centering
    \includegraphics[width=\linewidth]{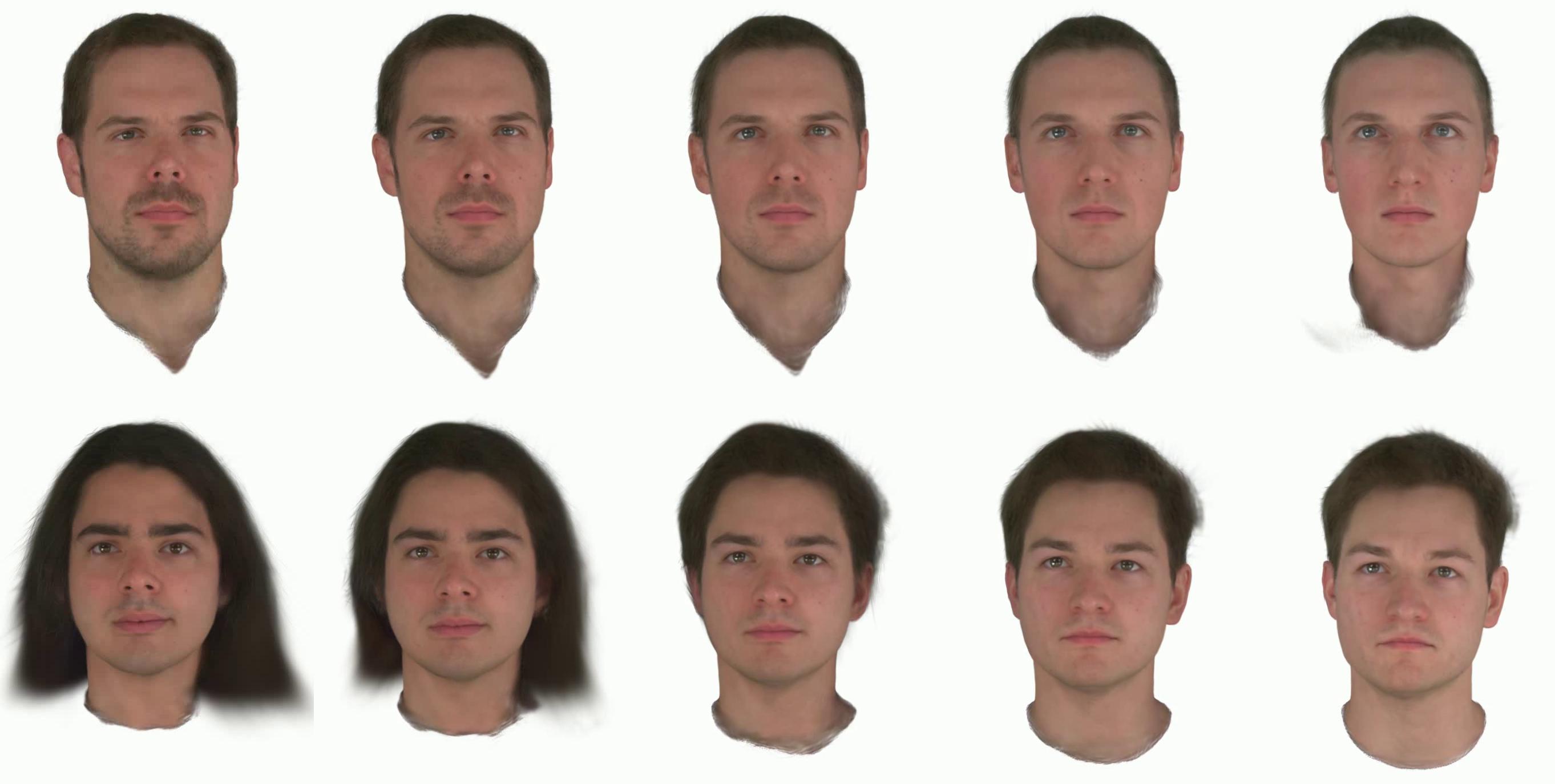}
    \caption{Interpolation between two test identities (leftmost and rightmost columns).}
    \label{fig:Interpolation}
\end{figure}
 Since, we introduce a different way of optimizing for the latent identity feature maps instead of a single latent code.
 Therefore, it is important to show that the latent feature space that we are optimizing is smooth.
 In this regard, we perform an experiment without enrollment for identity interpolation where we take two test identities and directly interpolate their latent feature maps after obtaining them from the identity encoder.
 We notice a smooth transition between the identities of our test subjects in Fig. \ref{fig:Interpolation}.
\\

\subsection{Cross-Reenactment}
\begin{figure}[t]
    \centering
    \includegraphics[width=\linewidth]{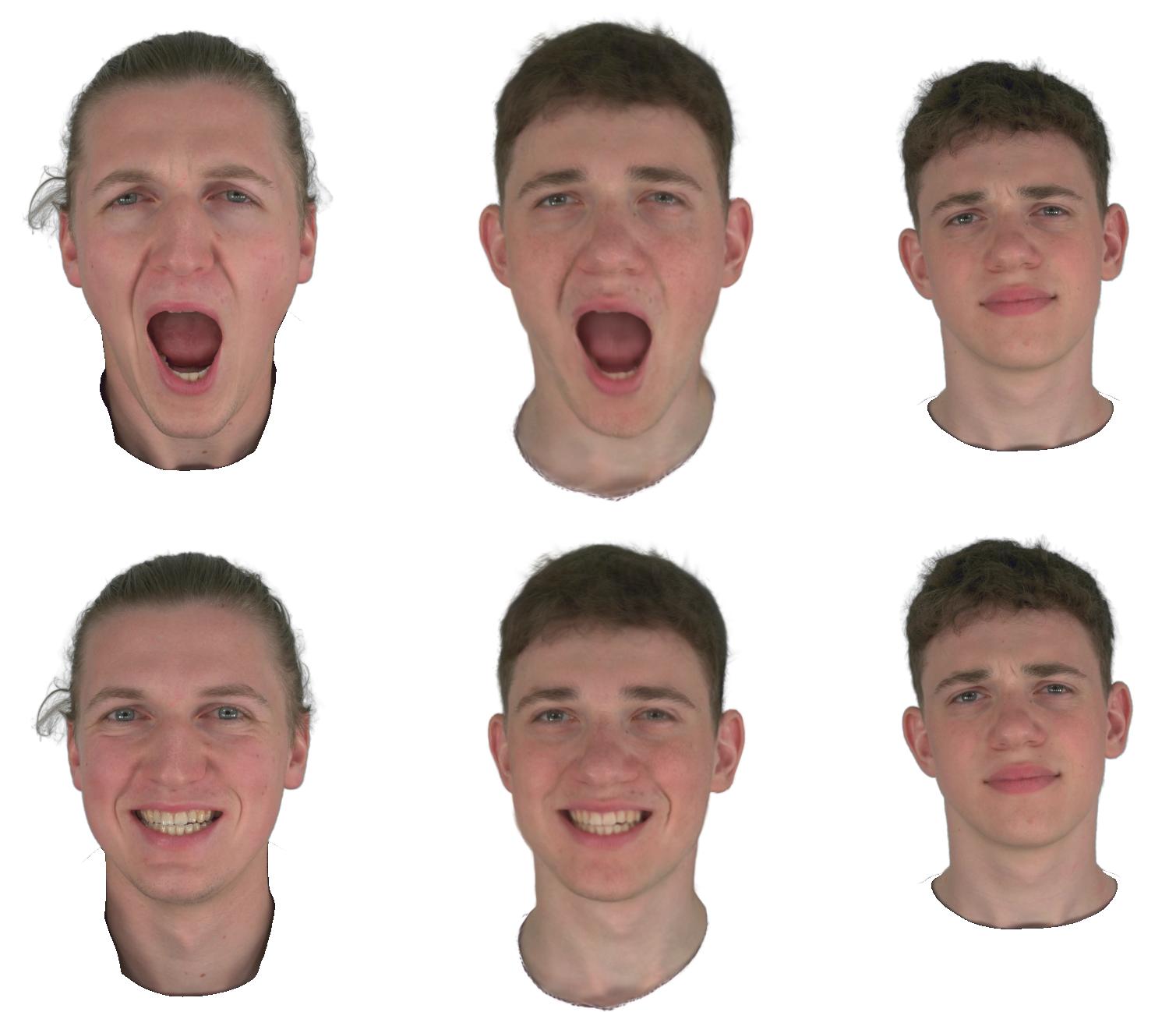}
    \caption{Cross-reenactment: driving actor (left), output (middle), identity of target (right).}
    \label{fig:Comparison_Cross_Reenactment}
\end{figure}

We also show additional experiments for cross-reenactment, even though it is not our main focus.
Our purpose here is just to demonstrate that our method gives plausible looking results for expression transfer between different identities, even though we notice that some part of the driving person's identity, particularly the structure of the mouth region is also transferred to the target actor as seen in Fig. \ref{fig:Comparison_Cross_Reenactment}.

\subsection{Inversion}
\begin{figure}[t!]
    \centering
    \includegraphics[width=\linewidth]{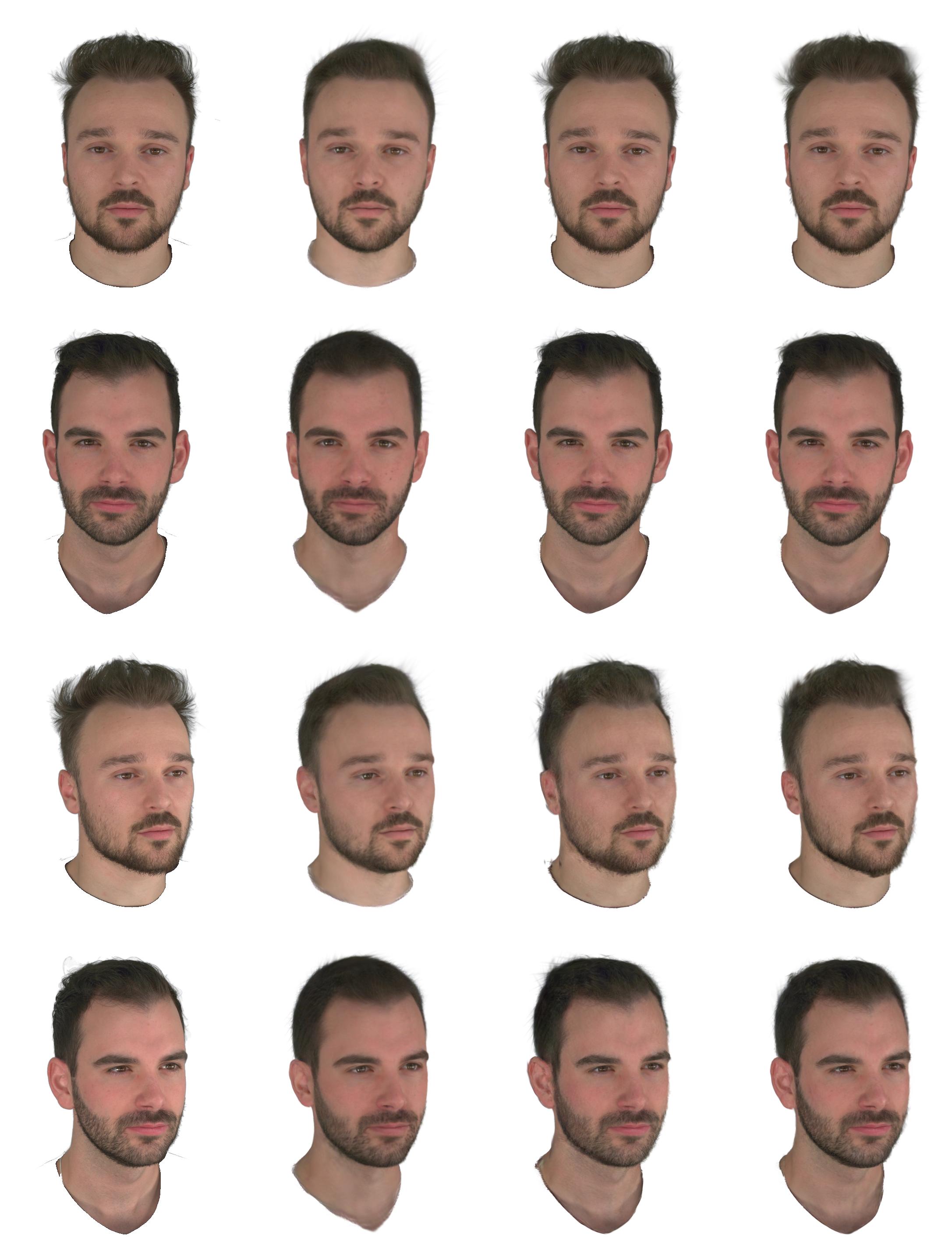}
    \footnotesize
    \begin{tabularx}{\linewidth}{YYYYYY} \\
        Ground Truth & No Fine-Tuning & 1-shot  & 3-shot
    \end{tabularx}
    \caption{Inversion Experiment for the No Fine-Tuning, 1-shot and 3-shot cases.}
    \label{fig:Inversion}
\end{figure}

We perform an experiment showcasing the effect of fine-tuning in general.
We compare a non-fine-tuning setting where we do not perform any optimization of the latent features against a given target during inference versus using a single or three images for inversion.
We can observe from Fig. \ref{fig:Inversion} that the quality as well as the identity consistency is better for the fine-tuning setting.
Moreover, for the side-views, we can clearly observe that the 3-shot results look sharper and consistent when rendered from the sides compared to non-fine-tuning and 1-shot cases.

\subsection{Comparison against GAN-based baselines}
\begin{figure}[t!]
    \centering
    \includegraphics[width=\linewidth]{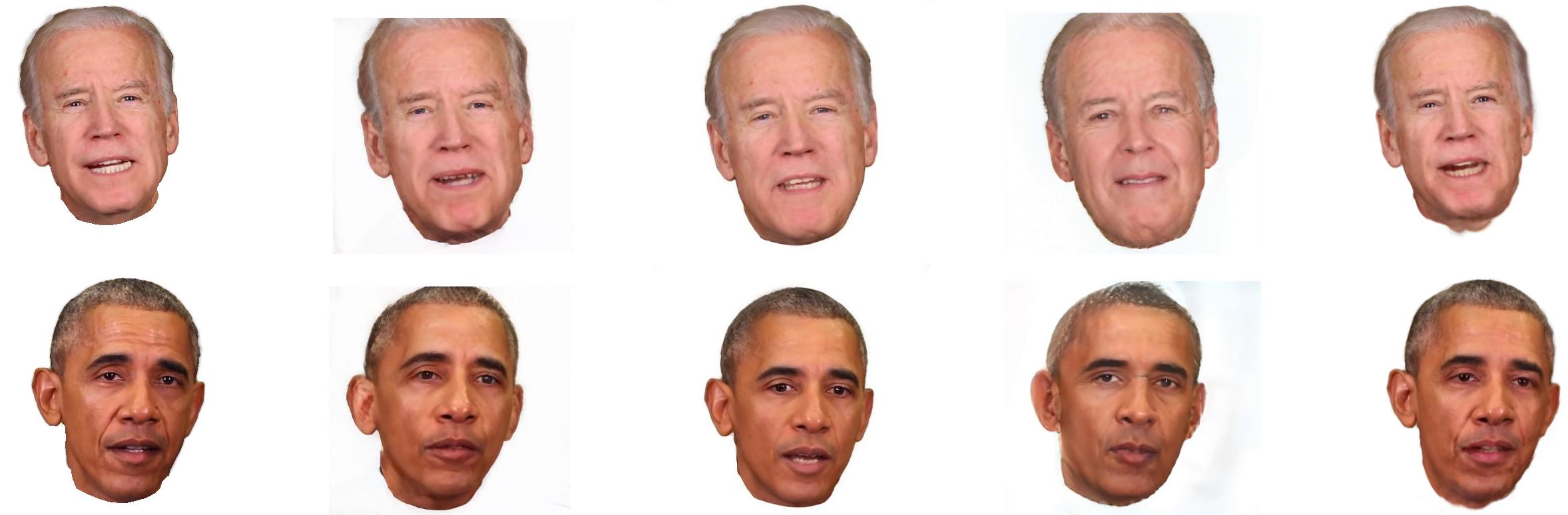}
    \footnotesize
    \begin{tabularx}{\linewidth}{YYYYY}
        Ground Truth & InvertAvatar \cite{invertavatar} & Portrait4D \cite{deng2024portrait4dlearningoneshot4d} &  Next3D \cite{sun2023next3d} & Ours
    \end{tabularx}
    \caption{Comparison against GAN-based baselines on In-the-Wild data.}
    \label{fig:CompGAN-based}
\end{figure}

We reuse the results from SynShot \cite{zielonka2025synshot} to further compare our method against GAN-based baselines namely InvertAvatar \cite{invertavatar}, Portrait4D \cite{deng2024portrait4dlearningoneshot4d} and  Next3D \cite{sun2023next3d} on In-the-Wild data from INSTA \cite{Zielonka2022InstantVH}. We can clearly deduce from Fig. \ref{fig:CompGAN-based} that our result looks closer to the ground truth driving frame compared to the other methods.

\subsection{Comparison against Avat3r}
\begin{figure}[t!]
    \centering
    \includegraphics[width=\linewidth]{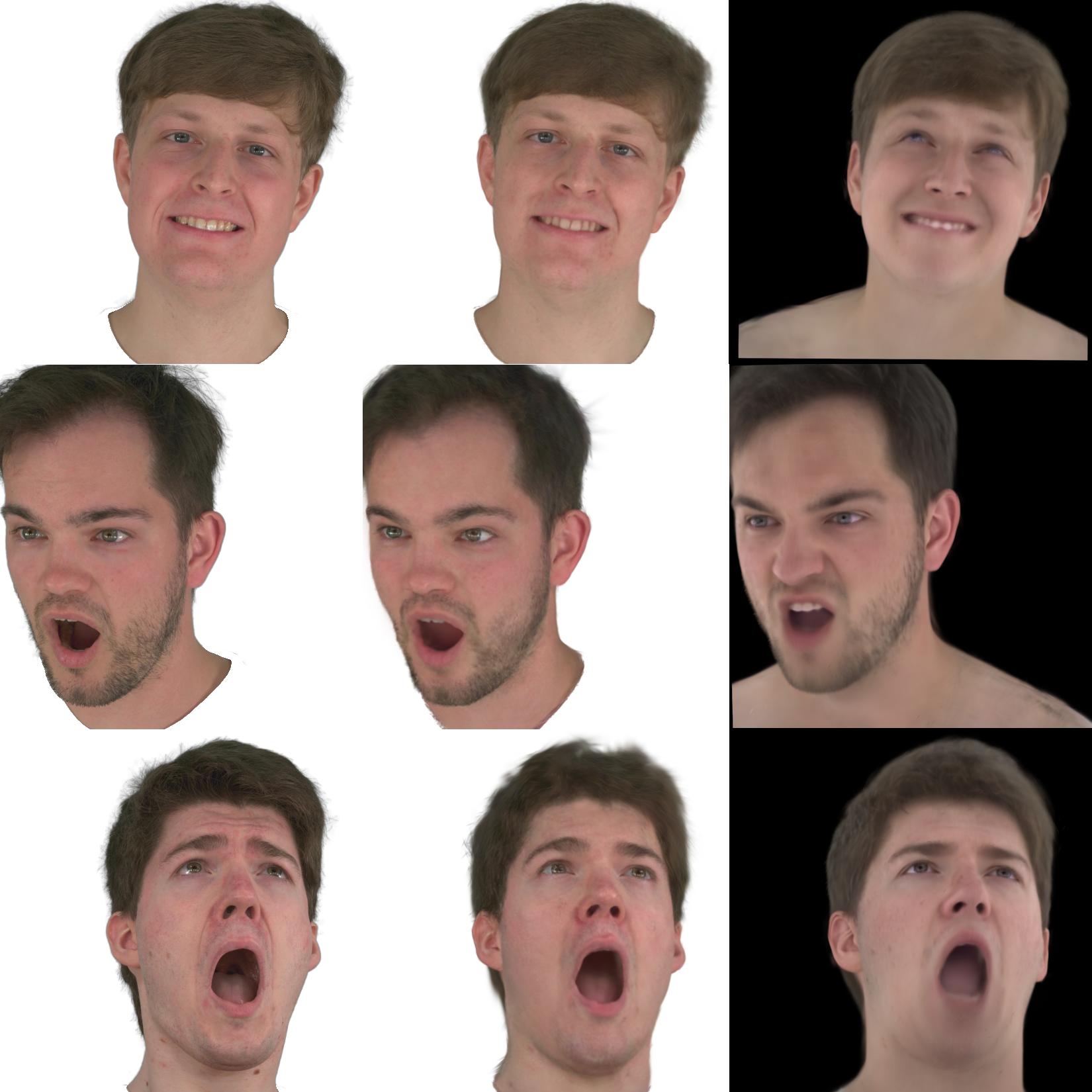}
    \footnotesize
    \begin{tabularx}{\linewidth}{YYYYYY}
        Ground Truth & Ours  & Avat3r~\cite{kirschstein2025avat3r}
    \end{tabularx}
    \caption{Comparison against Avat3r~\cite{kirschstein2025avat3r} for different subjects and driving frames rendered along different camera views.}
    \label{fig:Avat3r}
\end{figure}
We show qualitative comparison of our method against Avat3r \cite{kirschstein2025avat3r}, a prior-based method that operates in a multi-shot setting.
Since the code is not publicly available, we asked the authors to run some sequences for us.
The method takes as input 4 multi-view images of the actor's head and reconstructs a 3D avatar.
The avatars are then animated using expression codes and rendered along different camera views.
Our method, here, takes only 3 images during enrollment resulting in a 3D avatar and we render our avatar along the same camera views for the same driving frame for comparison.
We can see in Fig. \ref{fig:Avat3r} that the quality of our results is sharper compared to Avat3r and the latter also suffers from identity loss and fails to faithfully reproduce the driving expression.
\begin{figure}[t!]
    \centering
    \includegraphics[width=\linewidth]{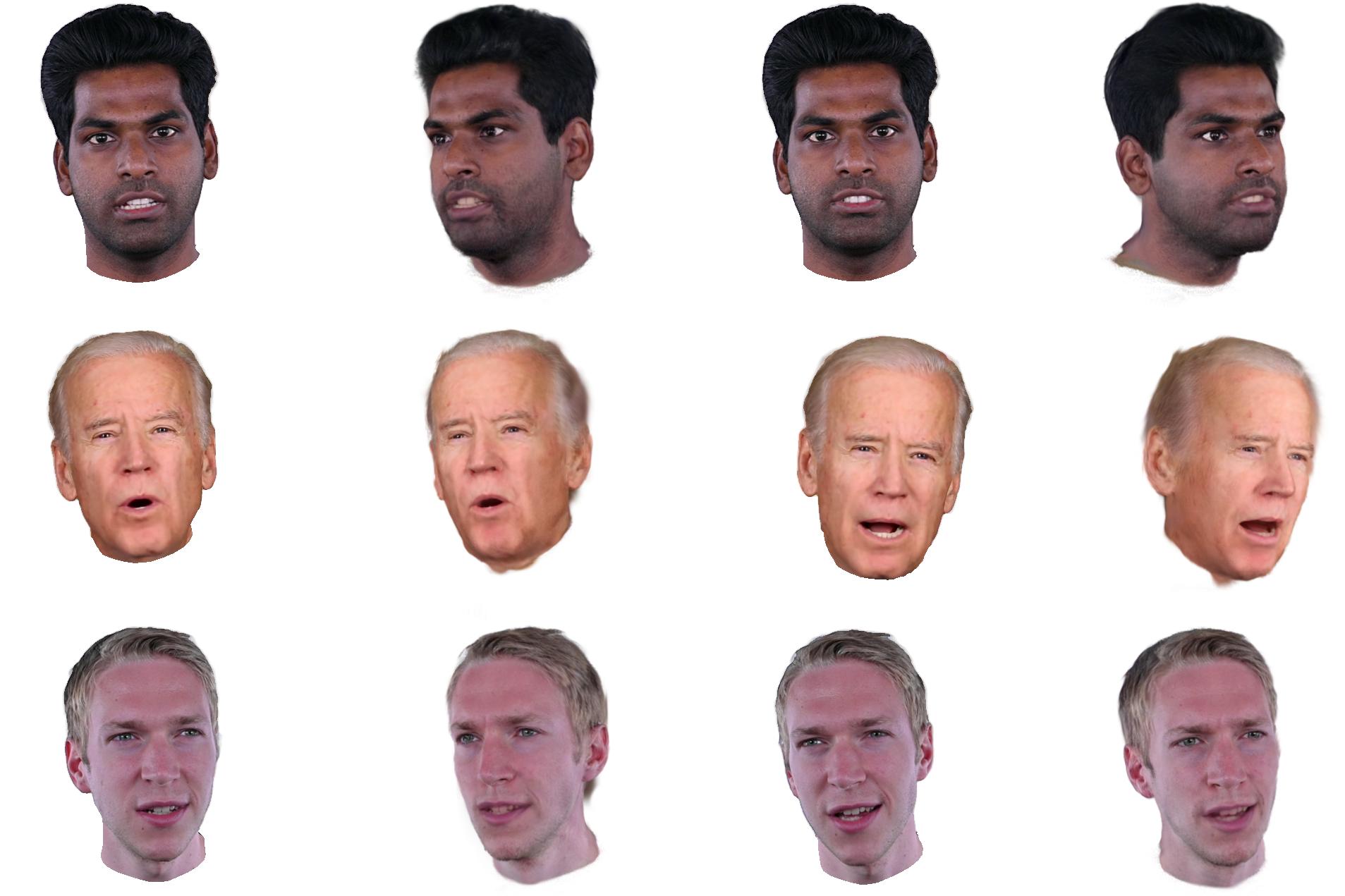}
    \footnotesize
    \begin{tabularx}{\linewidth}{YYYYY}
        Ground Truth & Novel View  &  Ground Truth & Novel View
    \end{tabularx}
    \caption{Novel View Synthesis on In-the-Wild data.}
    \label{fig:ITW_360}
\end{figure}

%


\subsection{In-the-Wild Novel View Synthesis}
In Fig. \ref{fig:ITW_360}, we show the novel side-views of In-the-Wild subjects. We can observe that our method gives plausible looking results when rendered from a novel view even though the target frames for some actors are strictly frontal-looking.

\subsection{Distribution of Gaussians}
\begin{figure}[t!]
    \centering
    \includegraphics[width=\linewidth]{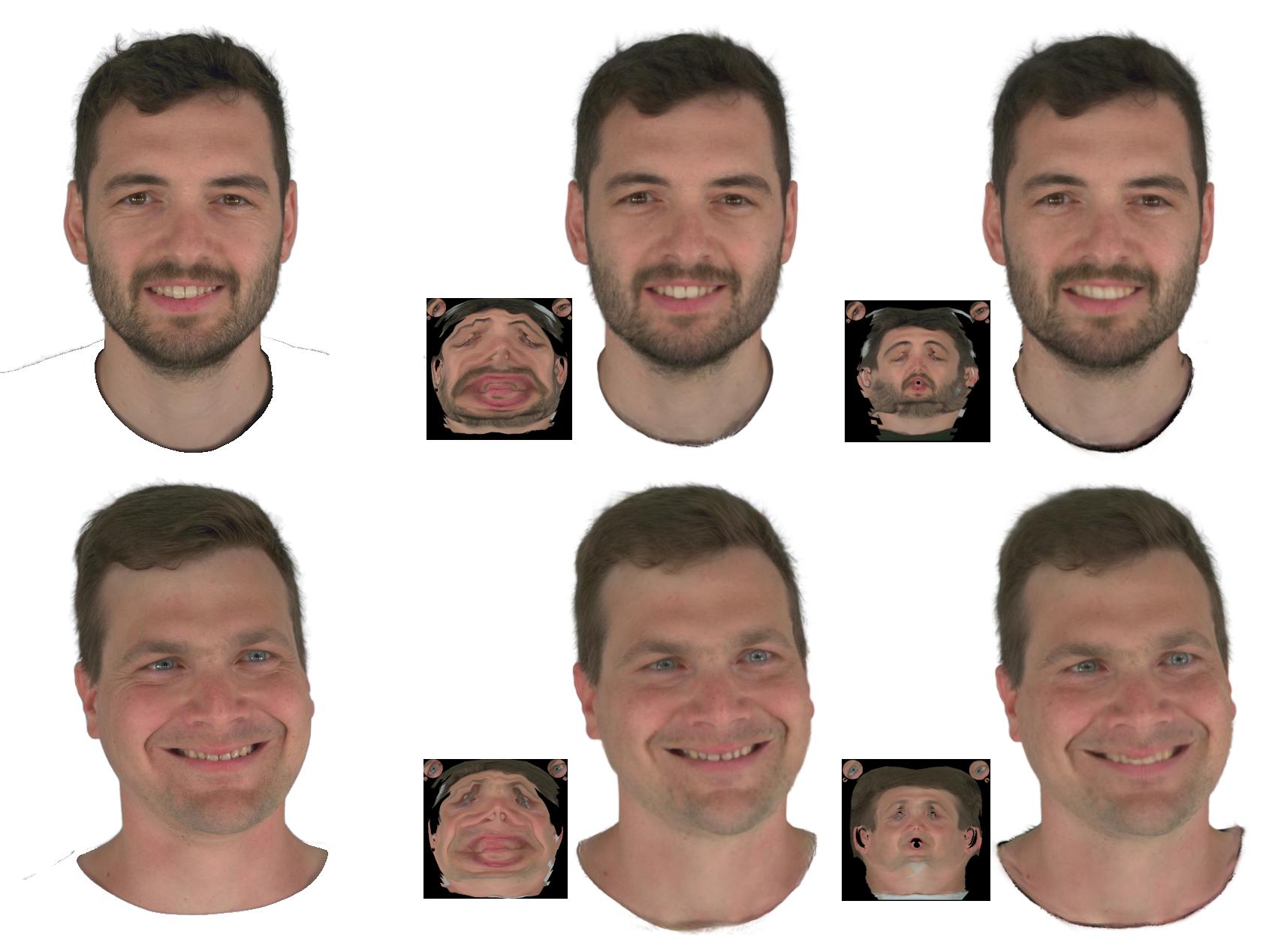}
    \footnotesize
    \begin{tabularx}{\linewidth}{YYYYYY}
        Ground Truth & Ours  & New UV
    \end{tabularx}
    \caption{Qualitative results of using a different UV map (shown on left side of each corresponding experiment) with different number and distribution of Gaussians in the mouth region.}
    \label{fig:NG}
\end{figure}

We perform an experiment showcasing the effect of varying the number of Gaussians around the mouth region. We use a different UV map than our current one. The new UV space is 10$\%$ smaller in terms of the total gaussian count but differs in the face region, particularly the mouth region with 70$\%$ lesser Gaussians. We can observe from Fig. \ref{fig:NG} and Tab. \ref{tab:ablation3} that even after significantly reducing the number of Gaussians, there is a slight degradation in quality compared to our UV space.
\subsection{3DMM-only control versus 3DMM with gradient maps}
\begin{figure}[t!]
    \centering
    \includegraphics[width=\linewidth]{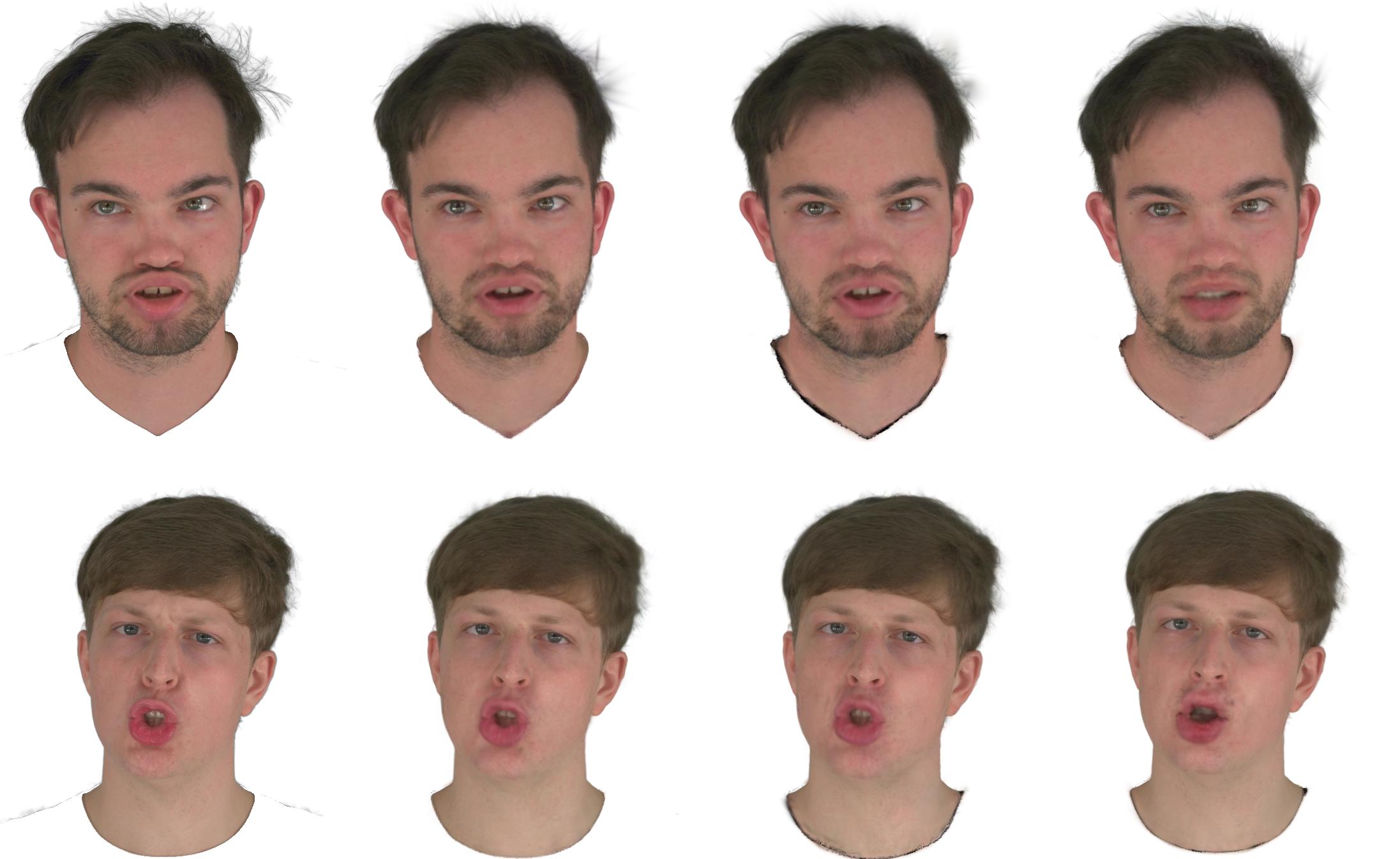}
    \footnotesize
    \begin{tabularx}{\linewidth}{YYYYYY}
        Ground Truth & Ours  & 3DMM-G & 3DMM-only
    \end{tabularx}
    \caption{Qualitative results of using 3DMM-only based control versus using additional mouth gradients (3DMM-G) versus our method (expression + mouth gradients).}
    \label{fig:3DMM}
\end{figure}

We conduct an ablation where we use only the underlying 3DMM-based control for our expressions without any additional expression map conditioning versus having an additional gradient-based mouth control signal. Note that we completely discard the expression-based conditioning in both cases. From Tab. \ref{tab:ablation4} and Fig. \ref{fig:3DMM}, we can see that even without the expression conditioning, the model with gradient-based control is much superior in reconstructing the mouth while the one with only a 3DMM-based control fails to reconstruct a meaningful mouth interior.
\subsection{Size of training set}
\begin{figure}[t!]
    \centering
    \includegraphics[width=\linewidth]{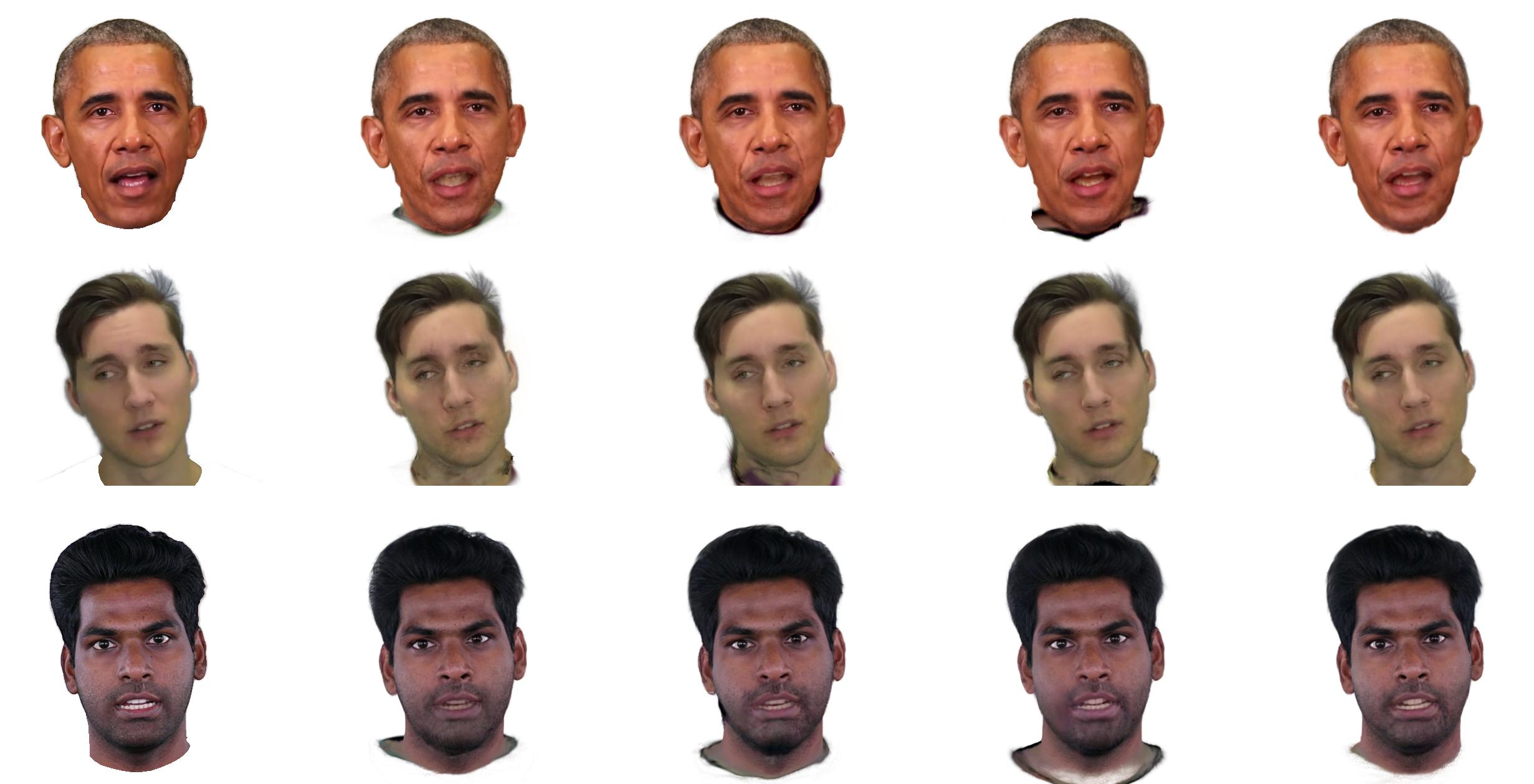}
    \footnotesize
    \begin{tabularx}{\linewidth}{YYYYYY}
        \vspace{1mm}
        Ground Truth & \vspace{1mm} 5$\%$ & \vspace{1mm} 25$\%$ & \vspace{1mm} 50$\%$ & \vspace{1mm} 100$\%$ (Ours)
    \end{tabularx}
    \caption{Qualitative comparison of different training set sizes.}
    \label{fig:Size_Img}
\end{figure}

We vary the size of training set and explore its impact on the test subject quality. We train our method with 50$\%$, 25$\%$ and 5$\%$ training data and compare against our model trained on full 100$\%$ data (200 subjects). We evaluate the models on In-the-Wild data. We show the comparison in Fig. \ref{fig:Size_Img} and Tab. \ref{tab:ablation5}, where we can see that having more training subjects improves quality but the improvement does not increase significantly as the size of training data increases. Surprisingly, the model gives plausible looking results even when trained on 5$\%$ data which we mainly attribute to color augmentations applied during training.

\subsection{Number of frames during Enrollment}
\begin{figure}[t!]
    \centering
    \includegraphics[width=\linewidth]{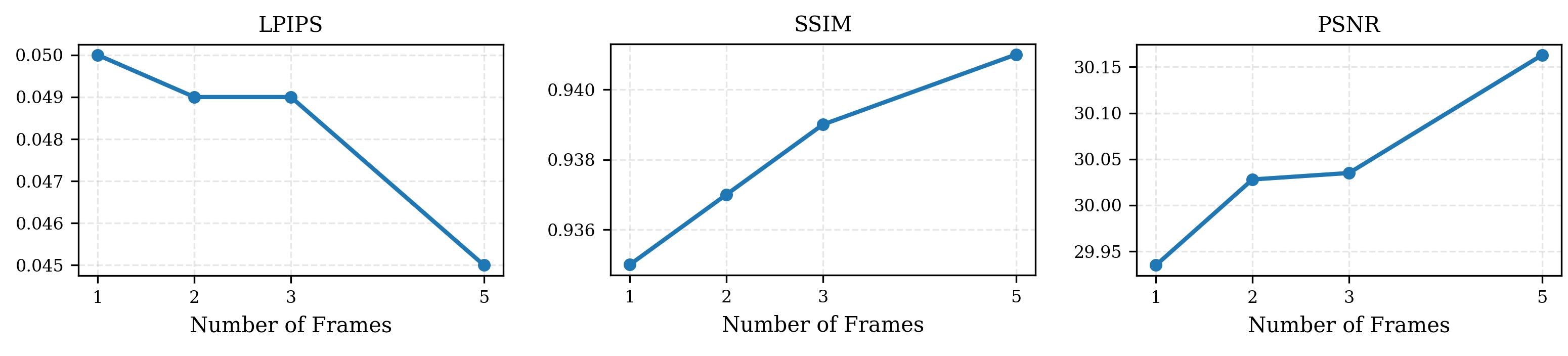}
    \caption{Plots of performance versus number of frames during inversion for LPIPS, SSIM and PSNR metrics.}
    \label{fig:Plots}
\end{figure}

\begin{figure}[!]
    \centering
    \includegraphics[width=\linewidth]{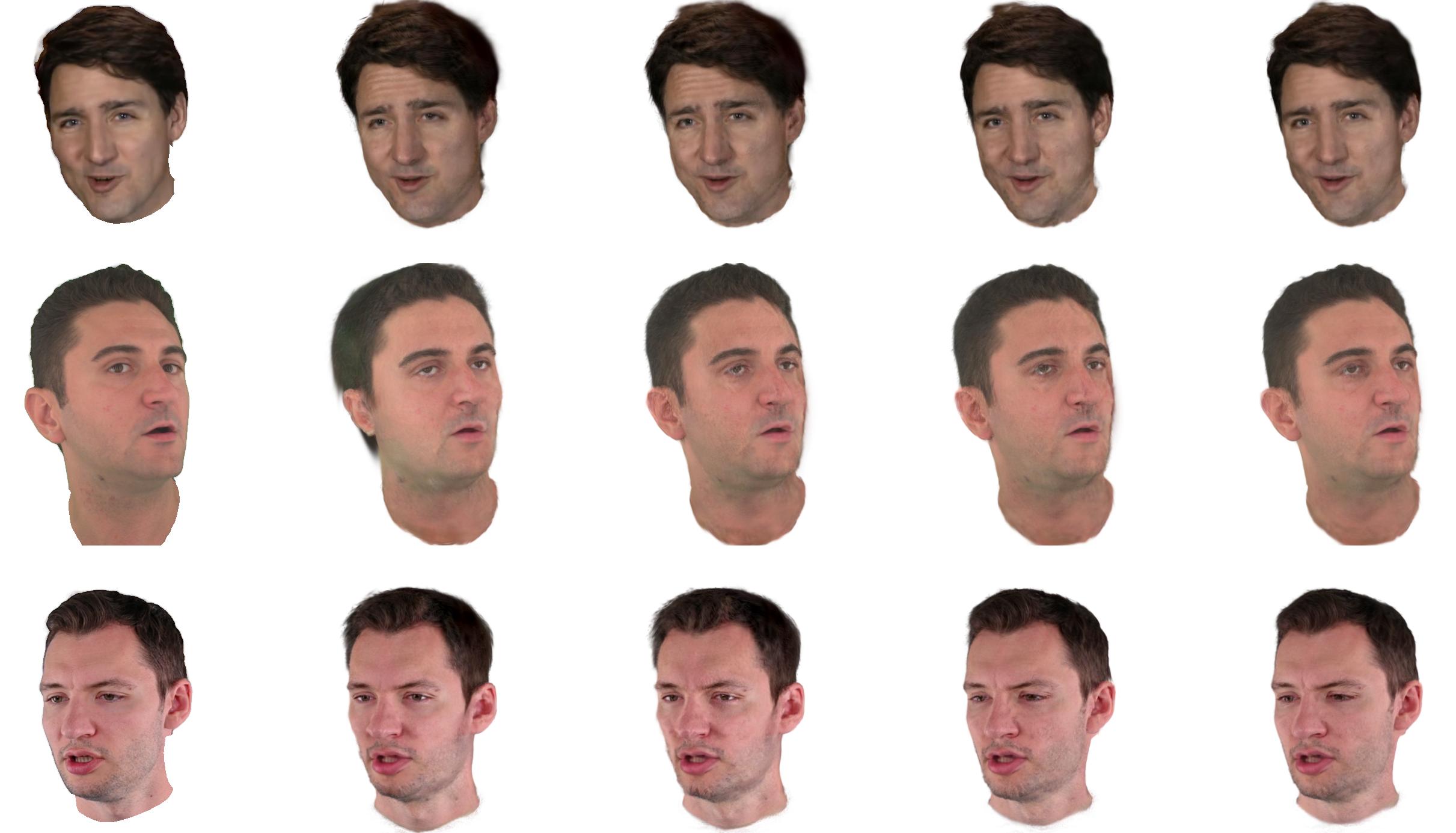}
    \footnotesize
    \begin{tabularx}{\linewidth}{YYYYYY}
    \vspace{1mm}
    Ground Truth & \vspace{1mm} 1 Frame & \vspace{1mm} 2 Frames & \vspace{1mm} 3 Frames & \vspace{1mm} 5 Frames
    \end{tabularx}
    \caption{Qualitative comparison between different number of frames used for inversion.}
    \label{fig:INV_Img}
\end{figure}

We perform an experiment where we vary the number of frames used for inversion during the enrollment stage. We start with a single frame followed by using 2, 3 and 5 frames covering the different views of In-the Wild subjects. We compute the metrics and plot them (LPIPS, SSIM and PSNR) in Fig. \ref{fig:Plots} to see the effect of this trade-off between quality and using more frames during enrollment. We can observe from Fig. \ref{fig:Plots}, Fig. \ref{fig:INV_Img} and Tab. \ref{tab:ablation6} that our method's performance improves with the number of frames.
\subsection{Comparison against GaussianAvatars}
\begin{figure}[!]
    \centering
    \includegraphics[width=\linewidth]{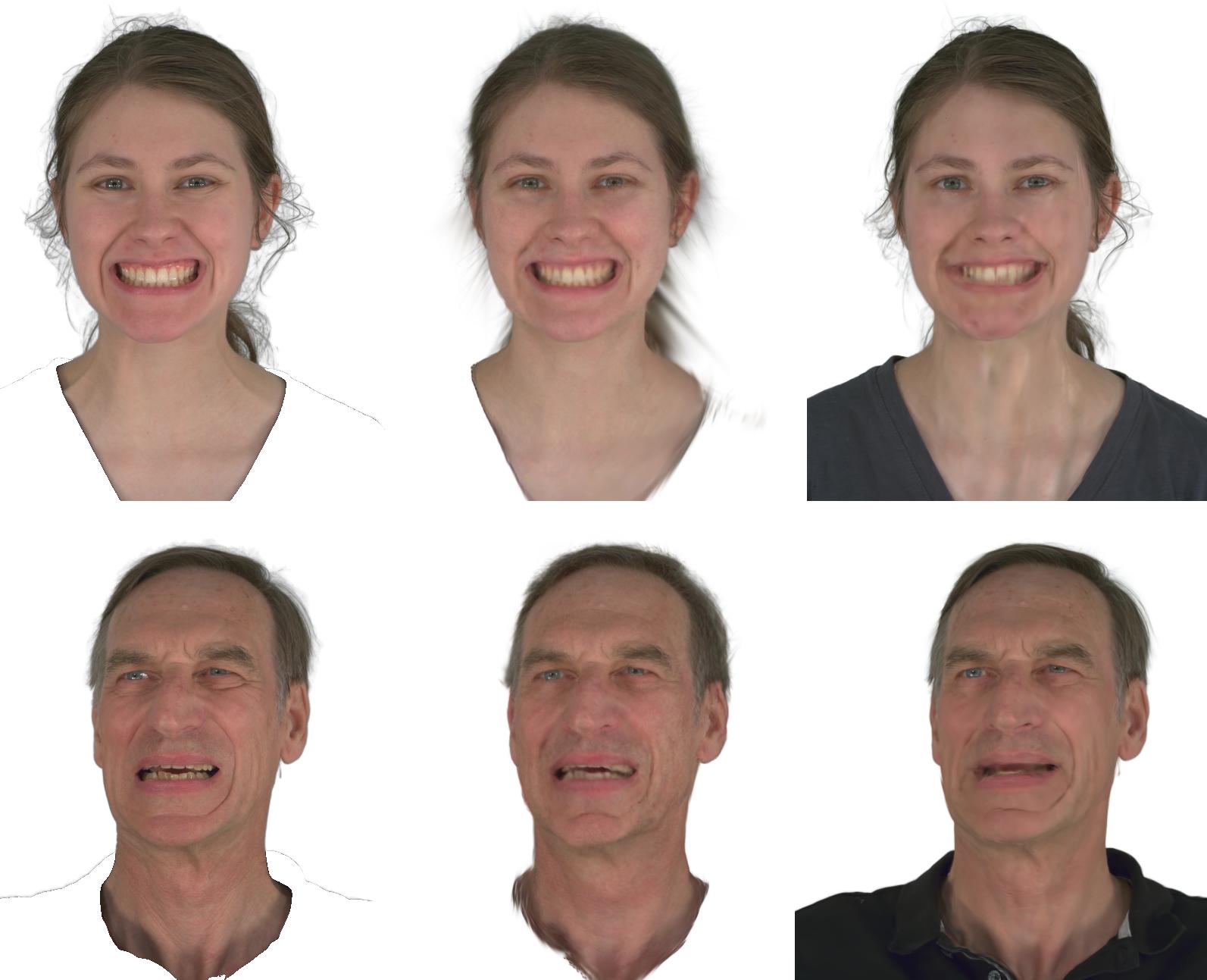}
    \footnotesize
    \begin{tabularx}{\linewidth}{YYYY}
    \vspace{1mm}
    Ground Truth & \vspace{1mm} Ours & \vspace{1mm} GaussianAvatars \cite{qian2024gaussianavatars}
    \end{tabularx}
    \caption{Qualitative comparison of our method against GaussianAvatars \cite{qian2024gaussianavatars}.}
    \label{fig:GA_Img}
\end{figure}

Finally, we perform a comparison against video-based method GaussianAvatars \cite{qian2024gaussianavatars}. We train the method on individual subject sequences and evaluate on the left-out frames of the expression sequences. We can observe from Fig. \ref{fig:GA_Img} that our method performs better, particularly in the mouth region.
\subsection{Category-wise evaluation}
We evaluate our method's performance on different categories of In-the-Wild subjects. Subject categories are manually annotated based on dominant attributes such as similarity to training data, less represented ethnicity and sequences with extreme views. While performance degrades mildly for more challenging out-of-distribution categories (\text{e.g.,} different ethnicity, extreme views), the model maintains relatively stable reconstruction quality and temporal consistency across all groups indicating generalization beyond the training distribution. From Tab. \ref{tab:ablation8}, we can see the metrics corresponding to subjects from each category.
\begin{table}[t]
    \caption{Quantitative evaluation of the distribution of mouth Gaussians on NeRSemble data.}
    \centering
    \small
    \resizebox{\linewidth}{!}{
    \begin{tabular}{|l|c|c|c|c|c|}
    \hline
    \textbf{Method} & \textbf{LPIPS} $\downarrow$ & \textbf{SSIM} $\uparrow$ & \textbf{PSNR} $\uparrow$ & \textbf{ID} $\uparrow$ & \textbf{t-LPIPS} $\downarrow$ \\
    \hline\hline
    Ours & 0.068$\pm$0.004 & 0.932$\pm$0.002 & 28.156$\pm$0.106 & 0.991$\pm$0.001 & 0.084$\pm$0.003 \\
    New UV (70$\%$ less) & 0.077 & 0.928 & 27.83 & 0.987 & 0.089 \\
    \hline
    \end{tabular}}
    \label{tab:ablation3}
    \vspace{-3.2mm}
\end{table}

\begin{table}[t]
    \caption{Quantitative evaluation of our method versus 3DMM-only control versus having both 3DMM control with additional Gradient-based mouth maps on NeRSemble data.}
    \centering
    \small
    \resizebox{\linewidth}{!}{
    \begin{tabular}{|l|c|c|c|c|c|}
    \hline
    \textbf{Method} & \textbf{LPIPS} $\downarrow$ & \textbf{SSIM} $\uparrow$ & \textbf{PSNR} $\uparrow$ & \textbf{ID} $\uparrow$ & \textbf{t-LPIPS} $\downarrow$ \\
    \hline\hline
    Ours & \textbf{0.068}$\pm$0.004 & \textbf{0.932}$\pm$0.002 & \textbf{28.156}$\pm$0.106 & \textbf{0.991}$\pm$0.001 & \textbf{0.084}$\pm$0.003 \\
    3DMM-only & 0.073 & 0.925 & 27.102 & 0.990 & 0.089 \\
    3DMM-Grad.  & \textbf{0.068} & \textbf{0.932} & 27.551 & \textbf{0.991} & 0.087 \\
    \hline
    \end{tabular}}
    \label{tab:ablation4}
    \vspace{-3.2mm}
\end{table}

\begin{table}[!t]
    \caption{Quantitative evaluation of the size of training set on 8 In-the-Wild subjects.}
    \centering
    \small
    \resizebox{\linewidth}{!}{
    \begin{tabular}{|l|c|c|c|c|c|}
    \hline
    \textbf{Training Set Size} & \textbf{LPIPS} $\downarrow$ & \textbf{SSIM} $\uparrow$ & \textbf{PSNR} $\uparrow$ & \textbf{ID} $\uparrow$ & \textbf{t-LPIPS} $\downarrow$ \\
    \hline\hline    
    100$\%$ (Ours) & \textbf{0.049} & \textbf{0.939} & \textbf{30.035} & \textbf{0.990} & \textbf{0.073} \\
    50$\%$  & \textbf{0.049} & 0.936 & 30.032 & 0.989 & 0.075 \\
    25$\%$  & 0.055 & 0.930 & 29.780 & 0.982 & 0.078 \\
    5$\%$ & 0.072 & 0.914 & 27.893 & 0.967 & 0.084 \\
    \hline
    \end{tabular}}
    \label{tab:ablation5}
    \vspace{-3.2mm}
\end{table}
\begin{table}[!t]
    \caption{Quantitative evaluation of the number of frames during enrollment on 8 In-the-Wild subjects.}
    \centering
    \small
    \resizebox{\linewidth}{!}{
    \begin{tabular}{|l|c|c|c|c|c|}
    \hline
    \textbf{No. of Frames} & \textbf{LPIPS} $\downarrow$ & \textbf{SSIM} $\uparrow$ & \textbf{PSNR} $\uparrow$ & \textbf{ID} $\uparrow$ & \textbf{t-LPIPS} $\downarrow$ \\
    \hline\hline
    5 Frames  & \textbf{0.045} & \textbf{0.941} & \textbf{30.163} & \textbf{0.992} & \textbf{0.071} \\
    3 Frames  & 0.049 & 0.939 & 30.035 & 0.990 & 0.073 \\
    2 Frames & 0.049 & 0.937 & 30.028 & 0.990 & 0.075 \\
    1 Frame & 0.050 & 0.935 & 29.935 & 0.987 & 0.076 \\
    \hline
    \end{tabular}}
    \label{tab:ablation6}
    \vspace{-3.2mm}
\end{table}
\begin{table}[!t]
    \caption{Category-wise quantitative evaluation of our method on 8 In-the-Wild subjects.}
    \centering
    \small
    \resizebox{\linewidth}{!}{
    \begin{tabular}{|l|c|c|c|c|c|c|}
    \hline
    \textbf{Subject Category} & \textbf{No. of Subjects} & \textbf{LPIPS} $\downarrow$ & \textbf{SSIM} $\uparrow$ & \textbf{PSNR} $\uparrow$ & \textbf{ID} $\uparrow$ & \textbf{t-LPIPS} $\downarrow$ \\
    \hline \hline
    Similar to training set & 3 & 0.050 & 0.930 & 30.20 & 0.989 & 0.071 \\
    Different ethnicity & 2  & 0.065 & 0.924 & 29.21 & 0.984 & 0.077 \\
    Extreme views & 3  & 0.054 & 0.929 & 29.22 & 0.989 & 0.075 \\
    \hline
    \end{tabular}}
    \label{tab:ablation8}
    \vspace{-3.2mm}
\end{table}
\end{document}